\documentclass[10 pt, journal]{IEEEtran}

\usepackage{amsmath, amssymb, amsthm}
\usepackage{hyperref, graphicx, cite, multirow}
\usepackage[caption = false, font = footnotesize]{subfig}
\usepackage{pgfplots, pgfplotstable}
\usepgfplotslibrary{external, statistics}
\usepackage{tikz}
\usetikzlibrary{shapes, arrows, positioning, plotmarks, patterns}

\pgfplotsset{colormap/bluered}

\definecolor{turquoise}{RGB}{0, 120, 200}

\hypersetup{
colorlinks = true,
linkcolor = black,
citecolor = black,
filecolor = black,
urlcolor = black,
bookmarks = true,
pdftoolbar = true,
pdfmenubar = true,
pdfstartview = {FitH},
pdftitle = {Sentinel: An Onboard Lane Change Advisory System for Intelligent Vehicles to Reduce Traffic Delay during Freeway Incidents},
pdfauthor = {Goodarz Mehr},
pdfsubject = {Diverge Advisory},
}

\begin{document}

	\title{Sentinel: An Onboard Lane Change Advisory System for Intelligent Vehicles to Reduce Traffic Delay during Freeway Incidents}
	
	\author{Goodarz Mehr and Azim Eskandarian, \IEEEmembership{Senior Member, IEEE}\thanks{Manuscript received September 15, 2020; revised April 1, 2021; accepted May 26, 2021. \textit{(Corresponding author: Goodarz Mehr.)}\\\indent The authors are with the Autonomous Systems and Intelligent Machines (ASIM) Laboratory, Virginia Tech, Blacksburg, VA 24061, USA. (email: \href{mailto:goodarzm@vt.edu}{goodarzm@vt.edu}; \href{mailto:eskandarian@vt.edu}{eskandarian@vt.edu}).\\\indent Digital Object Identifier: 10.1109/TITS.2021.3087578\\\indent$\copyright$ 2021 IEEE. Personal use of this material is permitted. Permission from IEEE must be obtained for all other uses, in any current or future media, including reprinting/republishing this material for advertising or promotional purposes, creating new collective works, for resale or redistribution to servers or lists, or reuse of any copyrighted component of this work in other works.}}
	
	\markboth{IEEE Transactions on Intelligent Transportation Systems}{Mehr \MakeLowercase{\textit{et. al.}}: Sentinel: An Onboard Lane Change Advisory System for Intelligent Vehicles to Reduce Traffic Delay during Freeway Incidents}
	
	\IEEEpubid{0000--0000/00\$00.00 \copyright\,2021 IEEE}
	
	\IEEEaftertitletext{\vspace{-4 pt}}
	
	\maketitle
	
	
	\begin{abstract}
		
		This paper introduces Sentinel, an onboard system for intelligent vehicles that guides their lane changing behavior during a freeway incident with the goal of reducing traffic congestion, capacity drop, and delay. When an incident blocking the lanes ahead is detected, Sentinel calculates the probability of leaving the blocked lane(s) before reaching the incident point at each time step. It advises the vehicle to leave the blocked lane(s) when that probability drops below a certain threshold, as the vehicle nears the congestion boundary. By doing this, Sentinel reduces the number of late-stage lane changes of vehicles in the blocked lane(s) trying to move to other lanes, and distributes those maneuvers upstream of the incident point. A simulation case study is conducted in which one lane of a four-lane section of the I-66 interstate highway in the U.S. is temporarily blocked due to an incident, to understand how Sentinel impacts traffic flow and how different parameters - traffic flow, system penetration rate, and incident duration - affect Sentinel's performance. The results show that Sentinel has a positive impact on traffic flow, reducing average delay by up to 37\%, particularly when it has a considerable penetration rate. Working alongside Traffic Incident Management Systems (TIMS), Sentinel can be a valuable asset for reducing traffic delay and potentially saving billions of dollars annually in costs associated with congestion caused by freeway incidents.
	
	\end{abstract}
	
	\section{Introduction} \label{Intro}
	
	\IEEEPARstart{F}{reeway} congestion can be categorized into two groups: recurrent congestion and non-recurrent congestion. Recurrent congestion is primarily due to traffic bottlenecks at fixed locations formed when traffic flow exceeds road capacity, for example at a lane drop. In contrast, non-recurrent congestion is generally caused by either freeway incidents (traffic accidents, vehicle breakdowns, etc.) or planned special events (freeway maintenance, construction, etc.) \cite{Ghosh}. The 2019 Urban Mobility Report published by the Texas Transportation Institute based on data obtained in 2017 found that through a combined 8.8 billion hours of wasted time and 3.3 billion gallons of wasted fuel, congestion costs the U.S. an estimated \$166 billion each year \cite{UrbanMob}. The National Traffic Incident Management Coalition (NTIMC) estimates that traffic incidents account for about 25\% of total delay on U.S. roadways \cite{NTIMC}. Therefore, any small improvement in traffic flow during freeway incidents can save a significant amount of time and money, and the emergence of intelligent vehicles presents new opportunities in this area. \par
	
	To reduce the impact of non-recurrent congestion - especially freeway incidents - on traffic flow, states and municipalities across the U.S. have traditionally relied on Traffic Incident Management Systems (TIMS). The goal of TIMS is to accurately predict the occurrence of traffic incidents and reduce the time for the detection of, response to, and clearance of an incident \cite{USFA}. Each of these aspects has been studied extensively by researchers in the field.
	
	\IEEEpubidadjcol
	
	Enhanced incident prediction allows TIMS to better prepare for them by allocating resources more efficiently, resulting in faster response and reduced delay. Past research in this area has investigated the use of parametric tree-based (CART or classification and regression tree) models, artificial neural networks, and negative binomial regression models for accident frequency prediction \cite{Chang1, Chang2}. Past works have also studied the use of random effects probit model for predicting freeway accident likelihood \cite{Qi}; using loop data through schematic eigenvectors for real-time crash likelihood prediction \cite{Xie}; and using the Hadoop framework to process and analyze big traffic data efficiently for accident prediction \cite{Park}. Similarly, automatic, fast, and accurate detection of traffic incidents with low false positive rates allows TIMS to respond faster and reduces the overall delay. Past studies in this area have employed fuzzy-wavelet radial basis function neural network \cite{Adeli, Karim}, constructive probabilistic neural network (CPNN) \cite{Jin2, Srinivasan}, wavelet transformation technique \cite{Teng}, and wavelet-clustering-neural network \cite{Ghosh} for freeway incident detection, proposing high accuracy incident detection methods with low false positive rates capable of adapting to different traffic conditions. \par
	
	Accurate estimation of incident duration also helps TIMS make more efficient decisions for responding to and clearing incidents, reducing their potential impact on traffic flow. Past studies in this area applied hazard-based duration models to statistically evaluate the time it takes to detect/report, respond to, and clear incidents \cite{Nam}, and used the survival analysis approach to develop a prediction model of accelerated failure time \cite{Junhua}. They also investigated a feature selection method that used genetic algorithms to create artificial neural network-based models that provide a sequential forecast of accident duration \cite{Lee}, and developed incident duration models for different incident types, studying several variables affecting incident duration \cite{Hojati}. More recent studies have proposed a novel M5P-HBDM model for accident duration prediction \cite{Lin} and gradient boosting decision trees (GBDTs) to predict the nonlinear and imbalanced incident clearance time based on different types of explanatory variables \cite{Ma}. They have also proposed a copula-based modeling framework for understanding the impact of influential factors on incident detection, response, and clearance times \cite{Zou} and a data-driven approach to automatically determine the spatiotemporal impact areas of freeway incidents \cite{Ou}. Overall, these studies have improved out understanding of the parameters that affect incident duration and enhanced the accuracy of estimating the time it takes to detect, respond to, and clear incidents. \par
	
	The final piece of the puzzle for TIMS is finding approaches for management of traffic upstream of the incident location during the time it takes to respond to and clear the incident from the road. To that end, TIMS have relied on approaches developed for managing recurrent congestion at road bottlenecks \cite{Farrag2}, including the use of variable speed limit (VSL) strategies \cite{Jin1, Yu1, Yu2}, congestion assistants \cite{VanDriel, Roncoli, Zhang1, Farrag1}, or a combination of both \cite{Zhang2, Farrag3}. Proposed VSL strategies generally involve dynamically altering a bottleneck's upstream speed limit through various optimization methods and control algorithms. Macroscopic simulations have shown these methods to be effective at reducing total travel time (TTT) and preventing capacity drop \cite{Yu2}; but as \cite{Zhang2} notes, lack of a lane assignment strategy can lead to the breakdown of these methods in microscopic simulations because delays and capacity drops often happen due to rushed lane changes close to the bottleneck. \par
	
	Even when a lane assignment strategy is present, deploying the traffic management approaches above relies on road infrastructure (e.g. overhead gantries) that are only available near large urban areas, limiting their effectiveness in dealing with traffic incidents that are random in nature. To address this problem, our research leverages the emergence and potential of connected, autonomous, and ADAS-enabled vehicles (here collectively referred to as intelligent vehicles) \cite{Zhou}. To that end, we introduce Sentinel, an onboard system for intelligent vehicles that utilizes a probabilistic prediction model \cite{Mehr1, Mehr2, Mehr3} to guide the lane changing behavior of individual vehicles during a freeway incident. Not only does this approach save significant amounts of time and money by reducing traffic delay during freeway incidents, it facilitates faster TIMS response and clearance of the road by improving traffic flow, reducing the chance of triggering secondary incidents \cite{Ma}. Furthermore, Sentinel only relies on a vehicle's onboard sensing capabilities and does not require any external road infrastructure, though its performance can be improved by leveraging connected vehicle technology. \par
	
	The remainder of this paper is organized as follows. \autoref{Method} presents the methodology, including a brief overview of the probability model underlying Sentinel, how Sentinel operates, and the simulation setup used to evaluate its performance. \autoref{Results} presents the results and discusses how Sentinel impacts traffic flow and efficiency. Finally, \autoref{Conclusions} summarizes the main findings of this paper and lays out the future path of this research.
	
	\section{Methodology} \label{Method}
	
	Before we can proceed with the methodology, we have to define a few terms. Throughout this work (including the previous section), a point on the road where upstream flow capacity is higher than downstream capacity is called a bottleneck \cite{Roncoli}. A bottleneck can be the result of road features such as merges and lane drops or temporary lane closures caused by incidents or road maintenance. Furthermore, as mentioned in \autoref{Intro} an intelligent vehicle is defined as a vehicle capable of achieving SAE Level 2 autonomy \cite{SAE}. In other words, the vehicle is equipped with one or more environment perception sensors (radar, camera, etc.) that enable the vehicle's advanced driver assistance systems (ADAS). \par
	
	Nominal capacity of a bottleneck is defined as the maximum traffic flow that can be maintained upstream and downstream of the bottleneck point simultaneously. In other words, if $C$ denotes the upstream capacity of a bottleneck point where a road transitions from $m$ lanes to $n$ lanes, $m > n$, the nominal bottleneck capacity $C_{b}$ is defined as $\frac{n}{m}C$. The actual bottleneck capacity, however, is often lower than the nominal capacity $C_{b}$, due to either upstream traffic flows larger than $C_{b}$ or rushed lane changing maneuvers of vehicles in the blocked lane(s) that disrupt traffic and slow other vehicles down. This reduction in capacity is called capacity drop and past research has shown that actual bottleneck capacity can be anywhere from 5\% to 20\% lower than the nominal capacity \cite{Cassidy, Chung}. Capacity drop disrupts traffic flow and results in increased delays for all vehicles. \par
	
	Sentinel, as introduced in \autoref{Intro}, is an onboard system utilizing perception data, knowledge of incident location, and a prediction model that estimates the likelihood of reaching a target point on the road using one or multiple lane changes \cite{Mehr1}, with the goal of reducing traffic delay and postponing or preventing capacity drop during freeway incidents. When an incident blocking the lane(s) ahead is detected, Sentinel calculates the probability of leaving the blocked lane(s) before reaching the incident point at each time step. It advises the vehicle to change lanes when that probability drops below a certain threshold, or in other words, as the vehicle nears the congestion boundary. By doing this, Sentinel reduces the number of late-stage lane changes of vehicles in the blocked lane(s) trying to move to other lanes and distributes those maneuvers upstream of the incident point. Detailed description of how Sentinel operates is provided in \autoref{SimSetup}, and information regarding our implementation can be found in \autoref{EDM}.
	
	\begin{figure*}[t!]
		\centering
		\includegraphics[width = \textwidth]{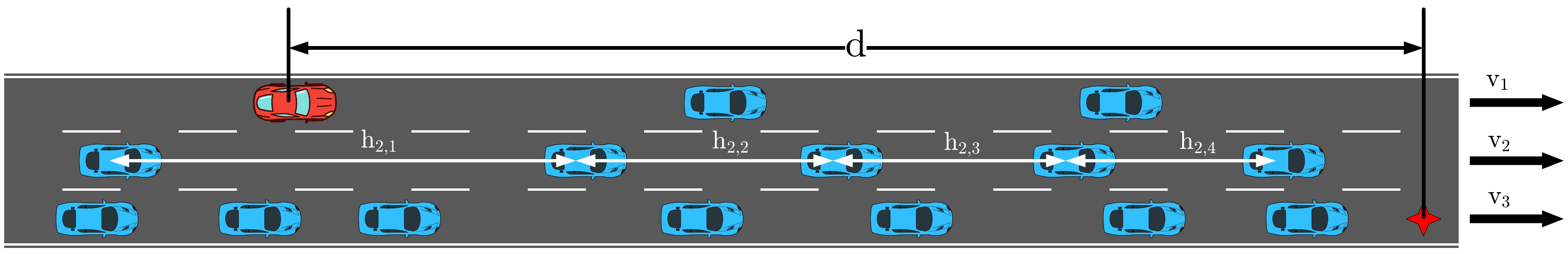}
		\caption{The prediction model estimates the probability that the red car (ego vehicle) can successfully change lanes to reach the red star (goal state) \cite{Mehr1}.} \label{ModelSchematic}
	\end{figure*}
	
	Sentinel was implemented in VISSIM\textsuperscript{\textregistered} for a series of simulations of a four-lane section of the I-66 interstate highway where the rightmost lane is blocked for a set amount of time. In what follows, \autoref{ProbModel} gives a brief overview of the probabilistic prediction model introduced above, while \autoref{SimSetup} describes the operation and implementation of Sentinel and the simulation setup used to evaluate its performance.
	
	\subsection{Probability model} \label{ProbModel}
	
	The model introduced in \cite{Mehr1} estimates the probability that a vehicle can reach a near-term goal state using one or multiple lane changes. While detailed model derivation and validation can be found in \cite{Mehr1}, a brief overview of the model is provided here to familiarize the reader. \par
	
	Consider a road with $n$ lanes, numbered by 1 to $n$ from left to right. Without loss of generality assume that the ego vehicle wants to reach a position on lane $n$ a distance $d$ ahead of its current position on lane 1 and let $P(S)$ denote the probability of doing so successfully. The model estimates this probability by making a few assumptions. First, it assumes that the velocity of all vehicles on lane $i$, $1 \le i \le n$, is equal to $v_{i}$, where $v_{i}$ can be considered the average velocity of all vehicles on that lane over a set period of time. Second, it assumes that inter-vehicle headway distances on lane $i$ are independent identically distributed (i.i.d.) random variables from a shared log-normal distribution defined by parameters $\mu_{i}$ and $\sigma_{i}$ \cite{Mei}. Finally, it assumes that the ego vehicle follows a Gipps gap acceptance model when changing lanes \cite{Gipps}. That is, if the ego vehicle is on lane $i - 1$, it only changes lanes if the gap between its leading and trailing vehicles on the adjacent lane $i$ is no smaller than a minimum acceptable (ciritical) gap $g_{i}$. This lane changing maneuver is completed in $t_{i}$ seconds. Some of these assumptions are shown in \autoref{ModelSchematic} for better visualization. \par
	
	The model calculates $P(S)$ based on the parameters defined above. In other words, for the case described above $P(S) = f_{n}(d, v_{1 : n}, \mu_{2 : n}, \sigma_{2 : n}, g_{2 : n}, t_{2 : n})$, where $q_{a : b}$ denotes $q_{a}, q_{a + 1}, \ldots, q_{b}$ for a parameter $q$ and indices $b \ge a$. $P(S)$ is obtained through induction on $n$ with $n = 2$ as the base case. For the base case, since $P(S)$ does not have a closed-form expression, it is obtained from a look-up table of values (as a function of three parameters) calculated using Monte Carlo simulations of the problem \cite{Mehr1, Mehr4}. For $n > 2$, $P(S)$ is determined recursively from
	\begin{equation} \label{ProbRecursive}
		\begin{split}
			&f_{n}(d, v_{1 : n}, \mu_{2 : n}, \sigma_{2 : n}, g_{2 : n}, t_{2 : n})\\
			&= \int_{0} ^ {d} f_{2}(d - x, v_{n - 1 : n}, \mu_{n}, \sigma_{n}, g_{n}, t_{n})\\
			&\times\frac{\partial}{\partial x} f_{n - 1}(x, v_{1 : n - 1}, \mu_{2 : n - 1}, \sigma_{2 : n - 1}, g_{2 : n - 1}, t_{2 : n - 1})\mathrm{d}x \\
			&= \frac{\partial}{\partial x}\int_{0} ^ {d} f_{2}(d - x, v_{n - 1 : n}, \mu_{n}, \sigma_{n}, g_{n}, t_{n})\\
			&\times f_{n - 1}(x, v_{1 : n - 1}, \mu_{2 : n - 1}, \sigma_{2 : n - 1}, g_{2 : n - 1}, t_{2 : n - 1})\mathrm{d}x,
		\end{split}
	\end{equation}
	which is based on the law of total probability \cite{Leon}. Extensive traffic simulations for a range of parameters confirmed that in the majority of cases the model is accurate to within 4\% of the actual probability. They also validated our observations that overall, the probability drops with decreased distance to the goal state (increased $d$), increased traffic density in adjacent lanes (increased $\mu_{i}$), or less aggressive driver behavior (increased $g_{i}$).
	
	\subsection{Simulation setup} \label{SimSetup}
	
	Sentinel leverages the probability model to guide the lane changing behavior of intelligent vehicles with the aim of reaching a particular goal state, here navigating the congestion caused by a freeway incident. Specifically, when a vehicle equipped with Sentinel approaches a freeway incident and is in the blocked lane(s), Sentinel uses knowledge of the location of that incident to calculate $d$ and utilizes perception and vehicle data to calculate $v_{i}, \mu_{i}$, and $\sigma_{i}$. Together with driver- or autonomous-vehicle-tuned parameters $g_{i}$ and $t_{i}$, at each time step Sentinel estimates the probability of the vehicle moving out of the blocked lane(s) (using one or more lane changes) under those conditions before reaching the incident point. It instructs the vehicle to change lanes when that probability drops below a certain threshold $p_{l}$. Using this approach, Sentinel reduces the number of late-stage, traffic disrupting lane changes of vehicles in the blocked lane(s) trying to get in other lanes and distributes those maneuvers upstream of the incident point. In other words, Sentinel's goal is to reduce congestion by changing the behavior of individual vehicles.
	
	\begin{figure*}[t!]
		\centering
		\includegraphics[width = \textwidth]{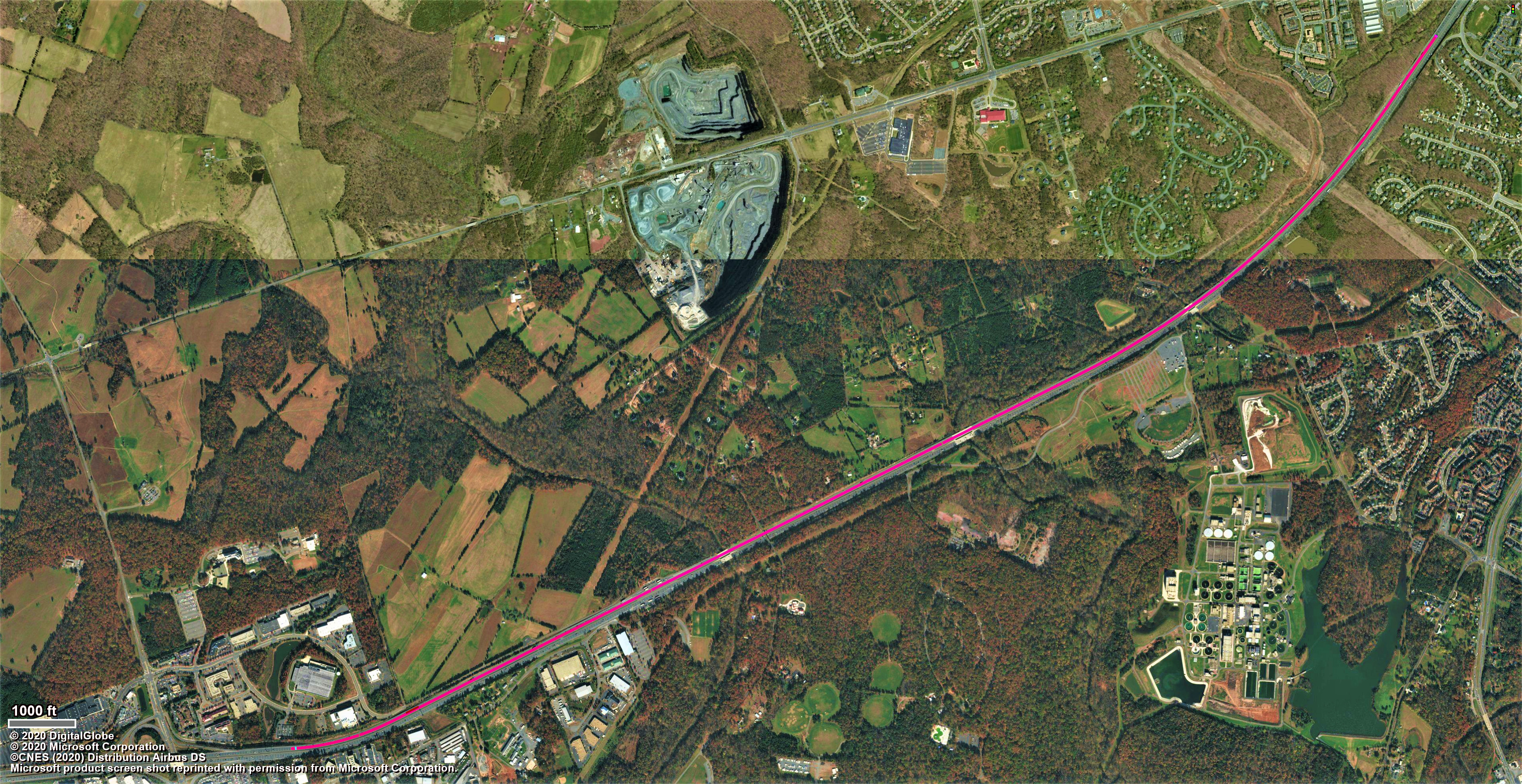}
		\setlength{\abovecaptionskip}{-10 pt}
		\caption{Satellite view of the westbound I-66 highway segment used for traffic simulations. Marked red in the photo, this segment has four lanes and is 3.99 mi long. It starts at the top right corner and ends at the bottom left corner.} \label{RoadSat}
	\end{figure*}
	\setlength{\abovecaptionskip}{0.5\baselineskip}
	
	Accurate knowledge of the incident is essential to the operation of Sentinel, as it is needed both to calculate $d$ and determine whether the vehicle is on a blocked lane. This information can be acquired from a few different sources. One source is TIMS, which as mentioned in \autoref{Intro} can detect incidents through a variety of methods \cite{Adeli, Karim, Jin2, Srinivasan, Teng, Ghosh}. Along with the location of the incident, TIMS can generally infer the number of blocked lanes by observing the discharge rate of vehicles downstream of the incident, though they cannot determine which exact lanes are blocked. A second source of information is crowdsourced incident and lane closure data provided by online navigation services like Waze and Google Maps. These services already allow users to report the location of traffic incidents, and it is possible that in the future users could also indicate which lanes have been blocked by an incident. The final mechanism for obtaining incident information is vehicle-to-vehicle (V2V) connectivity, through which a variety of perception, trajectory, and traffic information can be passed from one vehicle to another. Using this technology, a vehicle passing by an incident can report its location and the exact lanes blocked by it to vehicles upstream equipped with V2V communication technology. An intelligent vehicle could obtain and combine information from some or all of these three sources to determine the location of the incident and the lanes blocked by it \cite{Farrag1}. \par
	
	Sentinel would also need information about an intelligent vehicle's surrounding traffic to estimate the value of parameters such as $v_{i}, \mu_{i}$, and $\sigma_{i}$. The primary source of this information is the vehicle's own perception sensors. Most, if not all, intelligent vehicles are equipped with a few radar and camera sensors to enable ADAS such as lane keeping, adaptive cruise control (ACC), and emergency braking required for SAE Level 2 autonomy. Sentinel can tap into those sensor observations to determine the velocity of surrounding vehicles and their inter-vehicle headway distances, allowing it to estimate $v_{i}, \mu_{i}$, and $\sigma_{i}$. Of course, if vehicles are equipped with connected vehicle technology, V2V communication can be leveraged to obtain more accurate information about the velocity of surrounding vehicles and their inter-vehicle headway distances. \par
	
	As with any other ADAS, driver compliance is essential to the successful operation of Sentinel. For our case, non-compliant vehicles are those that do not change lanes according to Sentinel's advice, and are thus virtually indistinguishable from vehicles that are not equipped with Sentinel. Therefore, throughout this paper we assume vehicles equipped with Sentinel show full compliance and treat non-compliant vehicles and those not equipped with Sentinel as the same. \par
	
	VISSIM\textsuperscript{\textregistered} traffic simulations were used to study the impact of Sentinel on traffic flow, road capacity, and average delay during a freeway incident. Simulations were carried out for cases representing a variety of traffic conditions, obtained by changing incident duration, traffic flow, and Sentinel penetration rate. For each case, we investigated how different probability thresholds $p_{l}$ (the limiting value below which Sentinel advises the vehicle to leave the blocked lane(s)) impact road capacity, average delay, and overall traffic flow. \autoref{SimFund} to \autoref{DataProcessing} provide a detailed discussion of the simulation setup.
	
	\subsubsection{Simulation fundamentals} \label{SimFund}
	
	A segment of the westbound I-66 interstate highway on the outskirts of Washington, D.C. was selected for our simulations. As shown in \autoref{RoadSat}, the segment has four lanes and is 21,054 ft (3.99 mi) long. The starting point is immediately after the end of the acceleration lane of the merge from Lee Highway and the endpoint is just before the deceleration lane leading to the off-ramp to Sudley Road. The entire road segment is simulated as a single link with a vehicle input at the starting point. \par
	
	Traffic simulations were carried out according to Virginia Department of Transportation (VDOT) Traffic Operations and Safety Analysis Manual (TOSAM) and VDOT VISSIM\textsuperscript{\textregistered} User Guide \cite{TOSAM, VissimGuide}. A recommendation of \cite{VissimGuide} is to run each simulation case 10 times with different - but consistent - random seeds and average the results. However, we observed that for some cases a few simulation runs crashed before finishing\footnote{During a few runs, when Sentinel advised a vehicle to change lanes, VISSIM\textsuperscript{\textregistered}'s internal model that controlled the vehicle's driving behavior calculated a trajectory angle larger than 90 degrees, resulting in a crash.}, so we ran each simulation case 12 times\footnote{This number was limited by the available amount of RAM.}, starting from a random seed of 42 with increments of 5 for the following runs. Each simulation run lasted for 9000 seconds, with the first 1800 seconds as the seeding period and the rest as the analysis period \cite{VissimGuide}. \par
	
	A variety of simulation cases were studied based on different combinations of three parameters: input vehicle flow $q_{i}$, Sentinel penetration rate $r$, and incident duration $\gamma_{i}$. $q_{i}$ was set to either 6400, 7200, or 8000 veh/hr, $r$ was set to either 10\%, 40\%, or 70\%, and $\gamma_{i}$ was set to either 30 or 60 minutes. Out of the 18 possible cases, three where $\gamma_{i}$ = 60 min and $q_{i}$ = 8000 veh/hr were not studied because the resulting congestion would overflow the input. \par
	
	North American vehicles\footnote{Slightly larger than European vehicles used in VISSIM\textsuperscript{\textregistered} by default.} used in each simulation consisted of four different vehicle types: cars, Sentinel-equipped (SE) cars, buses (needed for modeling the incident), and heavy goods vehicles (HGVs). SE cars were identical to cars, with the only difference being the external driver model (EDM) that controlled their lane change initiation behavior, modeling Sentinel-equipped vehicles in real life (see \autoref{EDM} for details). Vehicle composition for SE cars, cars, HGVs, and buses was set according to $r$ in the following manner: $r$, 85\% - $r$, 13\%, and 2\%. For example, if $r$ = 70\%, the ratio of SE cars, cars, HGVs, and buses was set to 70\%, 15\%, 13\%, and 2\%, respectively. The desired speed distribution of all vehicles was set to 70 mph at the input, assigning a desired speed between 67 mph and 80 mph to each vehicle at random with uniform probability \cite{TOSAM}. Finally, a set of travel time measurements were defined to measure TTTs and delays throughout the simulation. Travel time was measured from the beginning of the road segment to near its end, covering a total distance of 21,000 ft. \par
	
	VISSIM\textsuperscript{\textregistered}'s public transport functionality was used to simulate a freeway incident. Specifically, a public transport station was defined on the rightmost lane at 19,000 ft with a total length of 200 ft. During each simulation, at 3600 seconds a bus would pull up to and stop at the station, effectively blocking traffic in that lane. The bus would leave the station at either 5400 seconds or 7200 seconds, modeling incidents where total incident time $\gamma_{i}$ was either 30 or 60 minutes.
	
	\subsubsection{Driving behavior} \label{DrivingBehavior}
	
	Data from a previous VDOT study \cite{Fairfax} was used in accordance with \cite{TOSAM, VissimGuide} to define the driving behavior for the single link. Parameter values shown in \autoref{DrivingBehaviorTable} were used due to anticipation of significant weaving and merging behavior induced by the incident.
	
	\begin{table}[t!]
		\renewcommand{\arraystretch}{1.2}
		\caption{Driving Behavior Parameters} \label{DrivingBehaviorTable}
		\centering
		\begin{tabular}{c c}
			\hline 
			Parameter & Value \\
			\hline
			CC0 (Standstill Distance) (ft) & 4.92 \\
			CC1 (Headway Time) (s) & 0.9 \\
			CC2 (Following Variation) (ft) & 13.12 \\
			\hline
			Maximum Deceleration (Own Vehicle) (ft/$\mathrm{s} ^ {2}$) & -15.00 \\
			Maximum Deceleration (Trailing Vehicle) (ft/$\mathrm{s} ^ {2}$) & -12.00 \\
			Accepted Deceleration (Own Vehicle) (ft/$\mathrm{s} ^ {2}$) & -4.00 \\
			Accepted Deceleration (Trailing Vehicle) (ft/$\mathrm{s} ^ {2}$) & -3.28 \\
			Safety Distance Reduction Factor & 0.25 \\
			Maximum Deceleration for Cooperative Braking (ft/$\mathrm{s} ^ {2}$) & -23.00 \\
			\hline
			Advanced Merging & On \\
			Cooperative Lane Change & On \\
			\hline
		\end{tabular}
	\end{table}
	
	\subsubsection{External driver model} \label{EDM}
	
	VISSIM\textsuperscript{\textregistered}'s external driver model (EDM) API provides access to various driving behavior aspects of all or different subsets of all vehicles. It was used in this work to model Sentinel's operation during a freeway incident. \par
	
	The first simulation of each case was run with all vehicles using VISSIM\textsuperscript{\textregistered}'s internal model, becoming the baseline for later comparison. In subsequent simulations, the EDM only affected the behavior of SE cars on the rightmost lane during the incident (i.e. between 3600 seconds and either 5400 or 7200 seconds, depending on incident duration). In the real world, Sentinel turns on when it is notified of an incident on the road ahead either by any of the methods described at the beginning of this section. \par
	
	To estimate the probability that a vehicle successfully changes lanes before reaching the incident point, Sentinel requires knowledge of $d, v_{i}, \mu_{i}, \sigma_{i}, g_{i}$, and $t_{i}$ for that vehicle. Because only the rightmost lane was blocked in our simulations, the EDM only needed $d, v_{1}, v_{2}, \mu_{2}, \sigma_{2}, g_{2}$, and $t_{2}$, assuming the rightmost lane was lane 1 and the lane to its left was lane 2. It set $d$ equal to the distance of the vehicle to the incident point and $v_{1}$ to the velocity of the vehicle. To estimate $v_{2}, \mu_{2}$, and $\sigma_{2}$, the EDM used data from vehicles in lane 2 up to 820.21 ft (250 m) or a maximum of 10 vehicles ahead and up to 492.13 ft (150 m) or a maximum of 2 vehicles behind the ego vehicle. This range is similar to the range of radars installed on most intelligent vehicles. Denoting by $m$ the number of detected vehicles and by $u_{i}$ and $r_{j}$, $1 \le i \le m$ and $1 \le j \le m - 1$, the velocities of these vehicles and their inter-vehicle headway distances, the EDM set
	\begin{align}
		v_{2} &= \bar{u} = \frac{\sum_{i = 1} ^ {m}u_{i}}{m}, \label{V2Bar} \\
		\mu_{2} &= \bar{\mu} =\frac{\sum_{j = 1} ^ {m - 1}\log(r_{j})}{m - 1}, \label{Mu2Bar} \\
		\sigma_{2} &= \bar{\sigma} = \mathrm{std}(\log(r_{j})), \label{Sig2Bar}
	\end{align}
	where std() is the standard deviation function. \autoref{Mu2Bar} and \autoref{Sig2Bar} come from the assumption that inter-vehicle headway distances are i.i.d. random variables from a shared log-normal distribution. \par
	
	The critical gap $g_{i}$ was set to $\delta v_{i} + s_{0}$ where $\delta$ and $s_{0}$ were set to 1.6 seconds and 1 meter, respectively. Though in reality the critical gap is stochastic in nature and depends on a multitude of factors including relative speeds and positions of nearby vehicles and driver aggressiveness, our choice simplified the model and its conservative nature (being generally larger than the actual critical gap) ensured lane changes were safe \cite{Toledo}. Finally, $t_{i}$ was set to 3 seconds because VISSIM\textsuperscript{\textregistered}'s internal model completes a lane change in that time \cite{PTV}. Both $g_{i}$ and $t_{i}$ can be tuned in a real-world implementation of Sentinel to match behavior characteristics of individual drivers or an autonomous vehicle. \par
	
	The only exception to this value assignment process came when $v_{2}$ was within the interval $v_{1} \pm v_{l}$, where $v_{l}$ was 4 m/s. In that case, we set $v_{2} = v_{1} + v_{l}$. The reasoning behind this goes back to \cite{Mehr1} that showed when $v_{1}$ is close to $v_{2}$, the relative traveled distance is significantly reduced, causing a large drop in probability that is unrealistic. Therefore, this modification was made to more accurately represent driver behavior during a lane changing maneuver. \par
	
	To simulate Sentinel, the EDM was tasked with calculating the probability described above at each time step. If the probability for a vehicle dropped below a certain threshold $p_{l}$, the EDM instructed that vehicle to change lanes. From \autoref{ProbModel} and intuition we know that the probability drops with decreased distance to the incident location and increased traffic density in adjacent lanes, so $p_{l}$ can be thought of as a measure for how far upstream of the congestion boundary Sentinel warns vehicles to change lanes. In other words, if $p_{l}$ is very high (close to 1), Sentinel is expected to warn vehicles far ahead of the congestion boundary, whereas if $p_{l}$ is low, Sentinel is expected to warn vehicles only when they near the congestion boundary. \par
	
	For each of the 15 cases, we tested different thresholds $p_{l}$ ranging from 0.999 to 0.6 to understand how $p_{l}$ affected average delay and traffic flow. Similar to \cite{Mehr2, Mehr3}, a problem we encountered during the implementation of Sentinel was that when the EDM signaled a vehicle to change lanes, that vehicle (driven by an internal VISSIM\textsuperscript{\textregistered} model) did not check to see if it was safe to do so, often passing through another vehicle during the maneuver. Our solution was to build a mechanism within the EDM to check for safety (adequate spacing and relative velocity of nearby vehicles) before starting a lane changing maneuver \cite{Toledo}. Details of the implementation of that mechanism can be found in \cite{Mehr2, Mehr3}.
	
	\subsubsection{Data processing and evaluation} \label{DataProcessing}
	
	Average delay, defined as the difference between travel time under free flow speed and actual travel time, was our measure of effectiveness (MoE) of choice \cite{VissimGuide}. Using the travel time measurement defined in \autoref{SimFund}, VISSIM\textsuperscript{\textregistered} calculated the delay for each vehicle. For each combination of $q_{i}, r$, $\gamma_{i}$, and $p_{l}$, we calculated the average $m_{i}$, standard deviation $s_{i}$, and maximum delay $a_{i}$ of all vehicles for each run $i$, $1 \le i \le 12$, reporting $m = \frac{1}{12}\sum_{i = 1} ^ {12} m_{i}, s = \frac{1}{12}\sum_{i = 1} ^ {12} s_{i}$, and $a = \frac{1}{12}\sum_{i = 1} ^ {12} a_{i}$ for each 900-second interval and the entire analysis period. \par
	
	As mentioned before, in some cases a few of the 12 runs crashed before finishing. Therefore, we utilized our approach in \cite{Mehr3}, excluding the crashed runs from $m, s$, and $a$ calculations for the analysis period but retaining data up to the nearest 900-second interval before the crash happened for calculations of $m, s$, and $a$ for 900-second intervals. For example, we retained data up to the 6300-second point if a run crashed at 6465 seconds. Cases with more than 2 crashes (hence fewer than 10 successful simulation runs) are marked with the number of successful runs in \autoref{30Min} and \autoref{60Min}.
	
	\section{Results and Discussion} \label{Results}
	
	\renewcommand{\tabcolsep}{2.8 pt}
	\begin{table*}[t!]
		\renewcommand{\arraystretch}{1.2}
		\caption{Statistical Indicators of Traffic Delay Results for $\gamma_{i}$ = 30 min} \label{30Min}
		\centering
		\begin{tabular}{| c | c | c c c | c c c | c c c |}
			\hline
			$q_{i}$ & \multirow{2}{*}{$p_{l}$} & \multicolumn{3}{c |}{$r$ = 10\%} & \multicolumn{3}{c |}{$r$ = 40\%} & \multicolumn{3}{c |}{$r$ = 70\%} \\
			\cline{3 - 11}
			(veh/hr) &  & Avg. (s) & Std. (s) & Max. (s) & Avg. (s) & Std. (s) & Max. (s) & Avg. (s) & Std. (s) & Max. (s) \\
			\hline
			\multirow{12}{*}{6400} & baseline & 18.1 & 32.9 & 522.9 & 18.1 & 32.9 & 522.9 & 18.1 & 32.9 & 522.9 \\
			 & 0.999 & 18.2 (\phantom{-}0.4) & 33.9 (\phantom{-}2.9) & 741.0 (41.7) & 16.1 (-11.0) & 28.3 (-14.1) & 497.5 (-\phantom{1}4.9) & 15.0 (-16.9) & 26.1 (-20.8) & 487.6 (-\phantom{1}6.7) \\
			 & 0.99 & 18.1 (\phantom{-}0.1) & 33.7 (\phantom{-}2.2) & 762.1 (45.7) & 16.0 (-11.6) & 28.2 (-14.5) & 454.8 (-13.0) & 14.7 (-18.9) & 25.4 (-23.0) & 487.9 (-\phantom{1}6.7) \\
			 & 0.97 & 17.8 (-1.6) & 32.7 (-0.6) & 705.8 (35.0) & 15.6 (-13.5) & 27.8 (-15.6) & 484.9 (-\phantom{1}7.3) & 14.9 (-17.7) & 25.4 (-22.8) & 433.2 (-17.2) \\
			 & 0.95 & 17.8 (-1.7) & 33.0 (\phantom{-}0.1) & 726.4 (38.9) & 16.4 (-\phantom{1}9.5) & 29.4 (-10.8) & 478.5 (-\phantom{1}8.5) & 14.3 (-20.8) & 24.4 (-26.0) & 458.4 (-12.3) \\
			 & 0.9 & 18.0 (-0.5) & 33.0 (\phantom{-}0.1) & 716.7 (37.1) & 16.1 (-10.8) & 28.7 (-12.9) & 505.4 (-\phantom{1}3.3) & 14.9 (-17.5) & 25.3 (-23.2) & 446.9 (-14.5) \\
			 & 0.85 & 17.6 (-2.5) & 32.4 (-1.7) & 682.2 (30.5) & 15.6 (-13.8) & 27.4 (-16.9) & 473.5 (-\phantom{1}9.4) & 15.1 (-16.8) & 25.8 (-21.7) & 446.4 (-14.6) \\
			 & 0.8 & 17.6 (-2.6) & 32.8 (-0.4) & 734.7 (40.5) & 16.2 (-10.7) & 28.9 (-12.2) & 485.6 (-\phantom{1}7.1) & 15.4 (-14.7) & 27.0 (-18.0) & 483.5 (-\phantom{1}7.5) \\
			 & 0.75 & 18.0 (-0.5) & 33.1 (\phantom{-}0.3) & 708.0 (35.4) & 16.7 (-\phantom{1}7.8) & 29.9 (-\phantom{1}9.3) & 510.8 (-\phantom{1}2.3) & 14.9 (-17.8) & 25.6 (-22.4) & 462.0 (-11.6) \\
			 & 0.7 & 17.7 (-2.0) & 33.0 (\phantom{-}0.2) & 760.9 (45.5) & 16.5 (-\phantom{1}8.8) & 29.2 (-11.5) & 481.4 (-\phantom{1}7.9) & 15.4 (-14.7) & 26.9 (-18.2) & 474.0 (-\phantom{1}9.4) \\
			 & 0.65 & 17.7 (-1.9) & 33.0 (\phantom{-}0.0) & 728.1 (39.2) & 16.2 (-10.3) & 28.9 (-12.4) & 487.1 (-\phantom{1}6.8) & 16.2 (-10.2) & 28.7 (-12.9) & 467.4 (-10.6) \\
			 & 0.6 & 18.2 (\phantom{-}0.7) & 33.7 (\phantom{-}2.2) & 740.5 (41.6) & 16.8 (-\phantom{1}7.0) & 30.5 (-\phantom{1}7.4) & 493.2 (-\phantom{1}5.7) & 15.3 (-15.1) & 26.7 (-19.0) & 446.0 (-14.7) \\
			\hline
			\multirow{12}{*}{7200} & baseline & 56.5 & 87.8 & 939.5 & 56.5 & 87.8 & 939.5 & 56.5 & 87.8 & 939.5 \\
			 & 0.999 & 55.5 (-1.7) & 88.0 (\phantom{-}0.2) & 805.4 (-14.3) & 51.4 (-\phantom{1}9.0) \phantom{$^{(9)}$} & 82.3 (-6.2) & 962.5 (\phantom{-}2.4) & 46.4 (-17.8) $^{(9)}$ & 76.9 (-12.4) & \phantom{1}876.7 (-\phantom{1}6.7) \\
			 & 0.99 & 55.5 (-1.8) & 87.2 (-0.7) & 740.2 (-21.2) & 49.6 (-12.1) \phantom{$^{(9)}$} & 79.7 (-9.2) & 925.6 (-1.5) & 46.9 (-16.9) $^{(6)}$ & 77.8 (-11.4) & \phantom{1}809.5 (-13.8) \\
			 & 0.97 & 54.5 (-3.5) & 85.5 (-2.6) & 749.0 (-20.3) & 49.9 (-11.6) \phantom{$^{(9)}$} & 80.7 (-8.0) & 902.8 (-3.9) & 45.4 (-19.6) \phantom{$^{(9)}$} & 75.7 (-13.8) & 1033.8 (\phantom{-}10.0) \\
			 & 0.95 & 55.2 (-2.2) & 86.6 (-1.3) & 763.9 (-18.7) & 51.8 (-\phantom{1}8.4) \phantom{$^{(9)}$} & 83.1 (-5.3) & 917.3 (-2.4) & 47.1 (-16.7) \phantom{$^{(9)}$} & 78.0 (-11.2) & \phantom{1}937.0 (-\phantom{1}0.3) \\
			 & 0.9 & 54.8 (-2.9) & 85.9 (-2.2) & 732.1 (-22.1) & 49.8 (-11.9) \phantom{$^{(9)}$} & 81.3 (-7.4) & 956.7 (\phantom{-}1.8) & 47.8 (-15.3) $^{(7)}$ & 79.3 (-\phantom{1}9.6) & 1209.2 (\phantom{-}28.7) \\
			 & 0.85 & 56.6 (\phantom{-}0.2) & 87.6 (-0.2) & 737.4 (-21.5) & 49.6 (-12.1) \phantom{$^{(9)}$} & 80.1 (-8.7) & 916.8 (-2.4) & 46.2 (-18.1) $^{(9)}$ & 76.7 (-12.6) & \phantom{1}934.0 (-\phantom{1}0.6) \\
			 & 0.8 & 55.6 (-1.5) & 86.3 (-1.6) & 730.7 (-22.2) & 51.2 (-\phantom{1}9.3) $^{(9)}$ & 81.8 (-6.9) & 865.7 (-7.9) & 48.0 (-14.9) $^{(9)}$ & 78.7 (-10.3) & 1099.8 (\phantom{-}17.1) \\
			 & 0.75 & 55.3 (-2.0) & 86.6 (-1.3) & 726.8 (-22.6) & 51.0 (-\phantom{1}9.7) \phantom{$^{(9)}$} & 81.7 (-6.9) & 911.4 (-3.0) & 47.1 (-16.7) $^{(9)}$ & 77.0 (-12.3) & \phantom{1}937.6 (-\phantom{1}0.2) \\
			 & 0.7 & 55.7 (-1.3) & 86.7 (-1.3) & 731.9 (-22.1) & 51.6 (-\phantom{1}8.6) \phantom{$^{(9)}$} & 83.2 (-5.2) & 944.4 (\phantom{-}0.5) & 48.0 (-14.9) \phantom{$^{(9)}$} & 79.4 (-\phantom{1}9.6) & \phantom{1}900.4 (-\phantom{1}4.2) \\
			 & 0.65 & 55.7 (-1.5) & 86.9 (-1.0) & 747.5 (-20.4) & 50.4 (-10.7) \phantom{$^{(9)}$} & 81.7 (-6.9) & 977.2 (\phantom{-}4.0) & 47.5 (-15.8) \phantom{$^{(9)}$} & 78.2 (-11.0) & \phantom{1}937.6 (-\phantom{1}0.2) \\
			 & 0.6 & 55.2 (-2.3) & 85.8 (-2.3) & 723.9 (-22.9) & 51.4 (-\phantom{1}8.9) \phantom{$^{(9)}$} & 83.3 (-5.1) & 941.8 (\phantom{-}0.2) & 48.5 (-14.1) \phantom{$^{(9)}$} & 79.7 (-\phantom{1}9.2) & 1020.5 (\phantom{-}\phantom{1}8.6) \\
			\hline
			\multirow{12}{*}{8000} & baseline & 118.5 & 140.0 & 1060.0 & 118.5 & 140.0 & 1060.0 & 118.5 & 140.0 & 1060.0 \\
			 & 0.999 & 113.7 (-4.0) & 140.6 (\phantom{-}0.5) & \phantom{1}965.1 (-9.0) & 110.4 (-6.8) \phantom{$^{(9)}$} & 137.0 (-2.1) & \phantom{1}932.8 (-12.0) & 102.8 (-13.3) \phantom{$^{(9)}$} & 130.6 (-6.7) & 895.8 (-15.5) \\
			 & 0.99 & 116.0 (-2.1) & 140.5 (\phantom{-}0.4) & \phantom{1}975.0 (-8.0) & 109.1 (-7.9) \phantom{$^{(9)}$} & 135.6 (-3.1) & \phantom{1}989.8 (-\phantom{1}6.6) & 101.7 (-14.1) $^{(7)}$ & 130.9 (-6.4) & 888.6 (-16.2) \\
			 & 0.97 & 117.8 (-0.6) & 139.1 (-0.6) & \phantom{1}978.3 (-7.7) & 110.2 (-7.0) $^{(7)}$ & 132.5 (-5.3) & 1014.1 (-\phantom{1}4.3) & 108.1 (-\phantom{1}8.7) $^{(4)}$ & 131.9 (-5.8) & 865.9 (-18.3) \\
			 & 0.95 & 116.8 (-1.4) & 140.2 (\phantom{-}0.2) & 1000.0 (-5.7) & 108.5 (-8.4) \phantom{$^{(9)}$} & 134.0 (-4.2) & \phantom{1}964.1 (-\phantom{1}9.0) & 102.5 (-13.5) $^{(7)}$ & 129.0 (-7.8) & 845.7 (-20.2) \\
			 & 0.9 & 119.0 (\phantom{-}0.4) & 141.2 (\phantom{-}0.9) & \phantom{1}981.9 (-7.4) & 108.4 (-8.5) $^{(7)}$ & 132.6 (-5.3) & \phantom{1}993.0 (-\phantom{1}6.3) & 102.6 (-13.4) $^{(9)}$ & 128.6 (-8.1) & 846.5 (-20.1) \\
			 & 0.85 & 115.1 (-2.9) & 140.4 (\phantom{-}0.3) & \phantom{1}992.1 (-6.4) & 109.7 (-7.4) \phantom{$^{(9)}$} & 134.6 (-3.8) & \phantom{1}935.7 (-11.7) & 104.8 (-11.6) $^{(8)}$ & 131.3 (-6.2) & 821.0 (-22.5) \\
			 & 0.8 & 117.1 (-1.2) & 141.6 (\phantom{-}1.1) & \phantom{1}961.5 (-9.3) & 110.1 (-7.1) $^{(9)}$ & 136.3 (-2.6) & \phantom{1}966.3 (-\phantom{1}8.8) & 107.7 (-\phantom{1}9.1) $^{(6)}$ & 136.7 (-2.4) & 933.6 (-11.9) \\
			 & 0.75 & 119.4 (\phantom{-}0.8) & 141.1 (\phantom{-}0.8) & \phantom{1}961.0 (-9.3) & 110.1 (-7.1) \phantom{$^{(9)}$} & 136.0 (-2.9) & 1000.3 (-\phantom{1}5.6) & 103.8 (-12.4) $^{(9)}$ & 133.0 (-5.0) & 889.5 (-16.1) \\
			 & 0.7 & 116.4 (-1.8) & 140.2 (\phantom{-}0.1) & \phantom{1}971.5 (-8.3) & 111.6 (-5.8) \phantom{$^{(9)}$} & 135.8 (-3.0) & 1015.1 (-\phantom{1}4.2) & 103.7 (-12.5) $^{(7)}$ & 133.3 (-4.8) & 852.7 (-19.6) \\
			 & 0.65 & 116.5 (-1.7) & 139.7 (-0.2) & \phantom{1}957.5 (-9.7) & 111.5 (-5.9) $^{(9)}$ & 136.9 (-2.2) & \phantom{1}961.3 (-\phantom{1}9.3) & 103.1 (-13.0) $^{(6)}$ & 129.4 (-7.5) & 799.5 (-24.6) \\
			 & 0.6 & 117.3 (-1.0) & 142.7 (\phantom{-}2.0) & \phantom{1}968.2 (-8.7) & 109.6 (-7.5) $^{(9)}$ & 137.1 (-2.0) & \phantom{1}984.9 (-\phantom{1}7.1) & 104.3 (-12.0) $^{(5)}$ & 132.1 (-5.6) & 840.0 (-20.8) \\
			\hline
		\end{tabular}
	\end{table*}
	\renewcommand{\tabcolsep}{6 pt}
	
	Statistical indicators of traffic delay results during the analysis period are shown in \autoref{30Min} for $\gamma_{i}$ = 30 min and in \autoref{60Min} for $\gamma_{i}$ = 60 min. In each table, the results are divided into square blocks based on the value of $q_{i}$ (increasing downward) and that of $r$ (increasing rightward). In each square block, each row provides the simulation results of that particular traffic setting (combination of $q_{i}, r$, and $\gamma_{i}$) for a particular value of $p_{l}$, with the first row being the baseline case and $p_{l}$ values decreasing downward from 0.999 in the second row. For each horizontal row of square blocks, $p_{l}$ values are provided in the second column. Furthermore, in each square block the columns represent average, standard deviation of, and maximum delay. The numbers in parenthesis next to delay values in each row - other than the baseline - indicate percentage change relative to the baseline case. For example, values in columns 9, 10, and 11 of the sixth row of \autoref{30Min} show the average, standard deviation of, and maximum delay for the simulation case with $q_{i}$ = 6400 veh/hr, $r$ = 70\%, $\gamma_{i}$ = 30 min, and $p_{l}$ = 0.97, with the numbers in parenthesis showing change relative to the baseline case in the third row. For this $p_{l}$ value, average delay was improved by 17.7\% relative to the baseline case, while standard deviation of the delay was improved by 22.8\%. Finally, as mentioned in \autoref{DataProcessing}, for instances where the number of crashed simulations exceeded 2, the number of simulations that finished without an error and were used for averaging the results is indicated as a superscript for the average delay value of that case and applies to all three values representing that case. For instance, for the case with $q_{i}$ = 7200 veh/hr, $r$ = 70\%, $\gamma_{i}$ = 30 min, and $p_{l}$ = 0.9, 5 of the 12 simulations crashed and the rest were used to calculate the results.
	
	\renewcommand{\tabcolsep}{2 pt}
	\begin{table*}[t!]
		\renewcommand{\arraystretch}{1.2}
		\caption{Statistical Indicators of Traffic Delay Results for $\gamma_{i}$ = 60 min} \label{60Min}
		\centering
		\resizebox{\textwidth}{!}{
		\begin{tabular}{| c | c | c c c | c c c | c c c |}
			\hline
			$q_{i}$ & \multirow{2}{*}{$p_{l}$} & \multicolumn{3}{c |}{$r$ = 10\%} & \multicolumn{3}{c |}{$r$ = 40\%} & \multicolumn{3}{c |}{$r$ = 70\%} \\
			\cline{3 - 11}
			(veh/hr) &  & Avg. (s) & Std. (s) & Max. (s) & Avg. (s) & Std. (s) & Max. (s) & Avg. (s) & Std. (s) & Max. (s) \\
			\hline
			\multirow{12}{*}{6400} & baseline & 42.8 & 65.5 & 714.9 & 42.8 & 65.5 & 714.9 & 42.8 & 65.5 & 714.9 \\
			 & 0.999 & 40.2 (-6.0) & 63.3 (-3.3) & 1153.3 (61.3) & 30.8 (-28.1) \phantom{$^{(9)}$} & 48.6 (-25.7) & 645.5 (-\phantom{1}9.7) & 27.8 (-35.0) \phantom{$^{(9)}$} & 44.5 (-32.0) & 620.6 (-13.2) \\
			 & 0.99 & 40.4 (-5.6) & 64.2 (-2.0) & 1194.4 (67.1) & 32.3 (-24.5) \phantom{$^{(9)}$} & 51.3 (-21.7) & 671.4 (-\phantom{1}6.1) & 28.1 (-34.5) $^{(9)}$ & 44.2 (-32.5) & 619.0 (-13.4) \\
			 & 0.97 & 39.3 (-8.1) & 62.1 (-5.1) & 1157.6 (61.9) & 30.6 (-28.5) \phantom{$^{(9)}$} & 48.2 (-26.4) & 637.2 (-10.9) & 26.8 (-37.4) \phantom{$^{(9)}$} & 42.3 (-35.5) & 586.2 (-18.0) \\
			 & 0.95 & 40.2 (-6.1) & 62.9 (-4.0) & 1143.3 (59.9) & 31.0 (-27.5) \phantom{$^{(9)}$} & 48.9 (-25.3) & 676.7 (-\phantom{1}5.3) & 27.2 (-36.4) $^{(9)}$ & 43.3 (-33.9) & 595.7 (-16.7) \\
			 & 0.9 & 40.3 (-5.8) & 62.3 (-4.8) & 1129.9 (58.1) & 33.7 (-21.3) \phantom{$^{(9)}$} & 52.8 (-19.4) & 622.3 (-13.0) & 29.1 (-32.1) \phantom{$^{(9)}$} & 46.0 (-29.8) & 596.1 (-16.6) \\
			 & 0.85 & 39.8 (-7.0) & 62.8 (-4.1) & 1127.0 (57.6) & 31.9 (-25.6) $^{(9)}$ & 50.2 (-23.4) & 626.1 (-12.4) & 28.5 (-33.4) \phantom{$^{(9)}$} & 45.3 (-30.8) & 607.7 (-15.0) \\
			 & 0.8 & 39.2 (-8.5) & 61.5 (-6.1) & 1173.2 (64.1) & 31.6 (-26.3) \phantom{$^{(9)}$} & 49.7 (-24.1) & 622.3 (-12.9) & 29.1 (-32.0) \phantom{$^{(9)}$} & 46.4 (-29.1) & 612.5 (-14.3) \\
			 & 0.75 & 41.1 (-4.0) & 64.3 (-1.9) & 1169.6 (63.6) & 34.3 (-19.9) \phantom{$^{(9)}$} & 53.6 (-18.2) & 672.6 (-\phantom{1}5.9) & 28.6 (-33.2) \phantom{$^{(9)}$} & 44.8 (-31.6) & 620.4 (-13.2) \\
			 & 0.7 & 39.7 (-7.2) & 63.0 (-3.8) & 1204.4 (68.5) & 35.0 (-18.2) \phantom{$^{(9)}$} & 54.3 (-17.0) & 659.4 (-\phantom{1}7.8) & 28.0 (-34.7) \phantom{$^{(9)}$} & 43.9 (-33.0) & 588.7 (-17.7) \\
			 & 0.65 & 39.8 (-7.0) & 63.3 (-3.3) & 1196.2 (67.3) & 33.2 (-22.5) \phantom{$^{(9)}$} & 52.4 (-20.0) & 637.4 (-10.8) & 31.0 (-27.6) $^{(9)}$ & 49.0 (-25.2) & 619.8 (-13.3) \\
			 & 0.6 & 40.6 (-5.3) & 63.8 (-2.5) & 1254.1 (75.4) & 35.4 (-17.3) \phantom{$^{(9)}$} & 56.0 (-14.5) & 667.4 (-\phantom{1}6.6) & 30.7 (-28.2) \phantom{$^{(9)}$} & 48.6 (-25.8) & 559.6 (-21.7) \\
			\hline
			\multirow{12}{*}{7200} & baseline & 185.2 & 180.9 & 1514.2 & 185.2 & 180.9 & 1514.2 & 185.2 & 180.9 & 1514.2 \\
			 & 0.999 & 178.4 (-3.7) \phantom{$^{(9)}$} & 178.0 (-1.6) & 1077.5 (-28.8) & 162.3 (-12.4) $^{(9)}$ & 163.4 (-\phantom{1}9.7) & 1611.5 (\phantom{-}\phantom{1}6.4) & 151.8 (-18.0) $^{(6)}$ & 159.8 (-11.7) & 1440.5 (-\phantom{1}4.9) \\
			 & 0.99 & 179.4 (-3.1) \phantom{$^{(9)}$} & 178.2 (-1.5) & 1104.4 (-27.1) & 157.0 (-15.2) $^{(7)}$ & 160.3 (-11.4) & 1047.4 (-30.8) & 149.5 (-19.3) $^{(4)}$ & 155.9 (-13.9) & 1049.8 (-30.7) \\
			 & 0.97 & 176.6 (-4.6) \phantom{$^{(9)}$} & 173.7 (-4.0) & 1051.4 (-30.6) & 163.9 (-11.5) \phantom{$^{(9)}$} & 167.5 (-\phantom{1}7.5) & 1570.5 (\phantom{-}\phantom{1}3.7) & 149.4 (-19.3) $^{(7)}$ & 155.7 (-13.9) & 1361.1 (-10.1) \\
			 & 0.95 & 180.7 (-2.4) \phantom{$^{(9)}$} & 179.2 (-1.0) & 1134.3 (-25.1) & 164.7 (-11.0) $^{(7)}$ & 167.9 (-\phantom{1}7.2) & 1316.3 (-13.1) & 149.0 (-19.5) $^{(4)}$ & 155.1 (-14.3) & \phantom{1}950.9 (-37.2) \\
			 & 0.9 & 176.7 (-4.6) $^{(9)}$ & 175.2 (-3.1) & 1082.3 (-28.5) & 162.6 (-12.2) $^{(8)}$ & 165.8 (-\phantom{1}8.4) & 1352.2 (-10.7) & 161.4 (-12.8) $^{(6)}$ & 166.5 (-\phantom{1}8.0) & 2307.8 (\phantom{-}52.4) \\
			 & 0.85 & 184.4 (-0.4) \phantom{$^{(9)}$} & 181.0 (\phantom{-}0.0) & 1082.5 (-28.5) & 166.7 (-10.0) $^{(6)}$ & 167.7 (-\phantom{1}7.3) & 1429.0 (-\phantom{1}5.6) & 143.9 (-22.3) $^{(2)}$ & 152.0 (-16.0) & \phantom{1}932.9 (-38.4) \\
			 & 0.8 & 181.0 (-2.3) \phantom{$^{(9)}$} & 179.5 (-0.8) & 1094.1 (-27.7) & 162.7 (-12.1) $^{(6)}$ & 166.7 (-\phantom{1}7.9) & 1463.1 (-\phantom{1}3.4) & 153.8 (-16.9) $^{(2)}$ & 155.0 (-14.3) & 1839.6 (\phantom{-}21.5) \\
			 & 0.75 & 179.0 (-3.4) \phantom{$^{(9)}$} & 178.3 (-1.5) & 1085.1 (-28.3) & 166.1 (-10.3) \phantom{$^{(9)}$} & 167.5 (-\phantom{1}7.4) & 1504.4 (-\phantom{1}0.6) & 153.8 (-16.9) $^{(6)}$ & 159.5 (-11.8) & 1470.8 (-\phantom{1}2.9) \\
			 & 0.7 & 182.1 (-1.7) \phantom{$^{(9)}$} & 178.5 (-1.4) & 1104.0 (-27.1) & 167.0 (-\phantom{1}9.8) \phantom{$^{(9)}$} & 169.2 (-\phantom{1}6.5) & 1580.5 (\phantom{-}\phantom{1}4.4) & 157.6 (-14.9) $^{(6)}$ & 163.5 (-\phantom{1}9.6) & 1448.7 (-\phantom{1}4.3) \\
			 & 0.65 & 180.0 (-2.8) \phantom{$^{(9)}$} & 176.4 (-2.5) & 1041.9 (-31.2) & 164.9 (-11.0) $^{(9)}$ & 166.0 (-\phantom{1}8.3) & 1629.0 (\phantom{-}\phantom{1}7.6) & 149.4 (-19.3) $^{(2)}$ & 157.2 (-13.1) & 2292.2 (\phantom{-}51.4) \\
			 & 0.6 & 181.4 (-2.1) \phantom{$^{(9)}$} & 177.3 (-2.0) & 1141.8 (-24.6) & 165.8 (-10.5) $^{(7)}$ & 167.7 (-\phantom{1}7.3) & 1005.4 (-33.6) & 155.6 (-16.0) $^{(4)}$ & 159.4 (-11.9) & 1660.7 (\phantom{-}\phantom{1}9.7) \\
			\hline
		\end{tabular}}
	\end{table*}
	\renewcommand{\tabcolsep}{6 pt}
	
	A first glance at the results shows that in all traffic settings (combinations of $q_{i}$, $r$, and $\gamma_{i}$), Sentinel was successful at improving traffic flow and reducing delay across all measures (average, standard deviation, and maximum) during the incident. The amount of reduction, however, depends on the parameters of each traffic setting and the chosen $p_{l}$ value. Therefore, we first discuss some overall trends in the results and then focus on the results of one simulation case and analyze them in more detail: the case with $q_{i}$ = 6400 veh/hr, $r$ = 40\%, and $\gamma_{i}$ = 60 min where all $p_{l}$ values result in around 20\% to 30\% reduction in average delay. \par
	
	\autoref{30Min} and \autoref{60Min} both show that delays for the baseline case are higher when incident duration $\gamma_{i}$ is longer, as expected. They also show that for longer incidents the reduction in delay is usually larger, especially for larger $r$ values. As will be discussed further ahead, this can be attributed to the cascading effect of delay reduction made possible by Sentinel. The results also show that average delay increases with $q_{i}$, because for larger traffic flows vehicles traveling in the blocked lane have a harder time finding an acceptable gap in the adjacent lane before getting close to the incident point, resulting in more last-minute lane changes that cause additional traffic disruptions and delays. Furthermore, the impact of the system is reduced with increasing $q_{i}$ because of congestion caused by a traffic flow beyond the capacity of a three-lane road. \par
	
	Sentinel was not active for baseline cases, so its penetration rate $r$ does not impact traffic delay results. When it was active, for $r$ = 10\% we see that it only had a modest impact on the results regardless of $q_{i}$ and $\gamma_{i}$ because of low penetration rate, matching our expectations. We also see that the reduction in delay increases with penetration rate $r$ and is largest for $r$ = 70\%, with improvements in average delay reaching as high as 37\% and showing that all vehicles reap the benefits of increased Sentinel penetration. As for $p_{l}$, no broad trends can be observed. Finally, we see that standard deviation and maximum values of delay follow similar trends as average delay for all parameters discussed above.
	
	\begin{figure}[t!]
		\centering
		\includegraphics[width = \columnwidth]{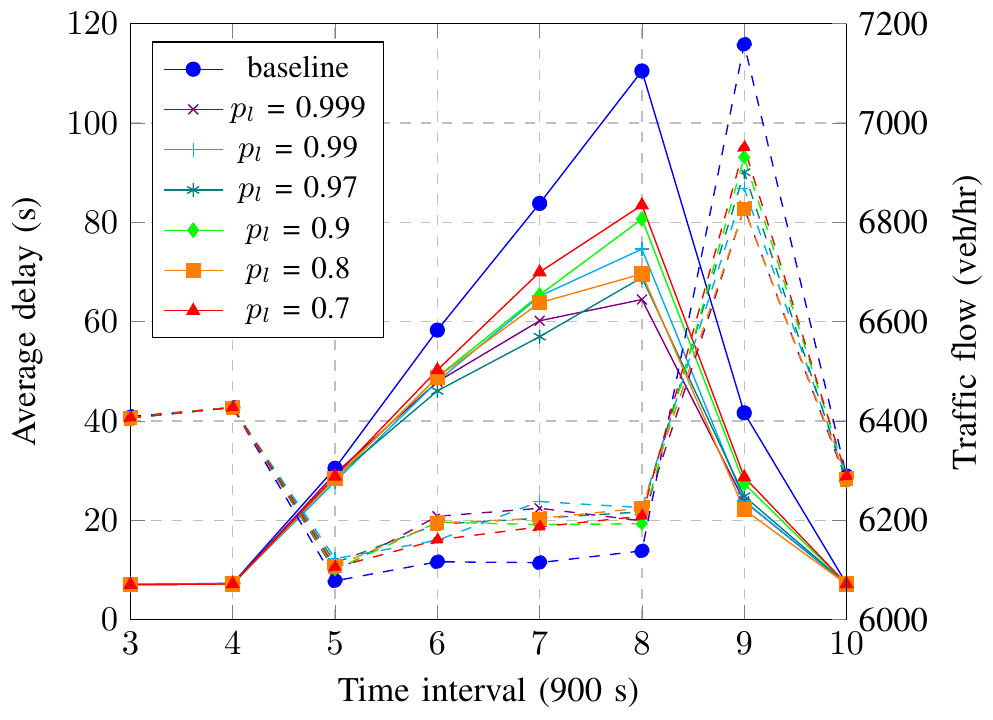}
		\caption{Average delay (solid lines) and discharge rate (dashed lines) of all vehicles for each 900-second simulation time-interval during the analysis period and for different values of $p_{l}$. The plot shows that overall, Sentinel is successful at reducing average delay over the analysis period and is able to restore some of the bottleneck capacity lost to congestion. It also shows that Sentinel's performance is dependent on the value of $p_{l}$.} \label{Case11All}
	\end{figure}
	\begin{figure}[t!]
		\centering
		\includegraphics[width = \columnwidth]{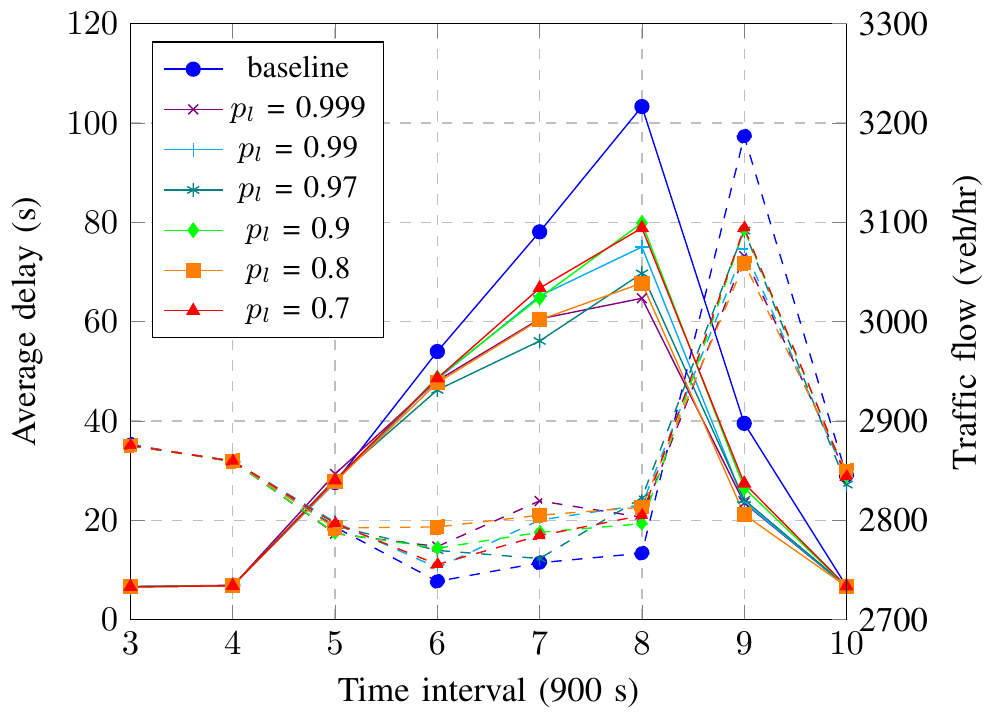}
		\caption{Average delay (solid lines) and discharge rate (dashed lines) of regular cars for each 900-second simulation time-interval during the analysis period and for different values of $p_{l}$.} \label{Case11Regular}
	\end{figure}
	
	For the next analysis we need to define a new parameter $d_{l}$, known as lane departure density \cite{Mehr3}. $d_{l}$ can be thought of as a measure for quantifying the time-space rate of vehicles departing a blocked lane for the last time before reaching the blockage point. More specifically, if $N$ vehicles depart a blocked lane for the last time within a certain time interval with duration $T$ and road span with length $D$ before the blockage point, $d_{l}$ for that $T$-$D$ time-space block is defined as
	\begin{equation} \label{LDDensity}
		d_{l} = \frac{N}{DT}.
	\end{equation}
	$d_{l}$ has the units of (lane departure)/(ft.s). In this paper we assume that $T$ = 100 s and $D$ = 200 ft.
	
	\begin{figure}[t!]
		\centering
		\includegraphics[width = \columnwidth]{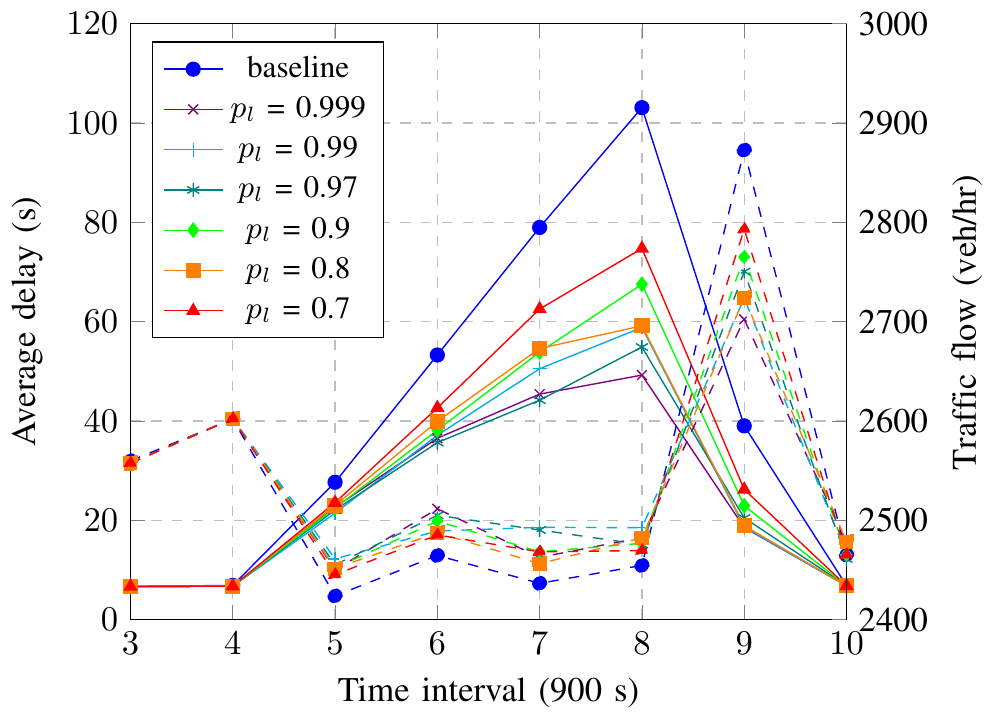}
		\caption{Average delay (solid lines) and discharge rate (dashed lines) of SE cars for each 900-second simulation time-interval during the analysis period and for different values of $p_{l}$. The plot shows that compared to regular cars, SE cars experience less delay on average in cases other than baseline.} \label{Case11Smart}
	\end{figure}
	
\begin{figure}[t!]
		\centering
		\includegraphics[width = \columnwidth]{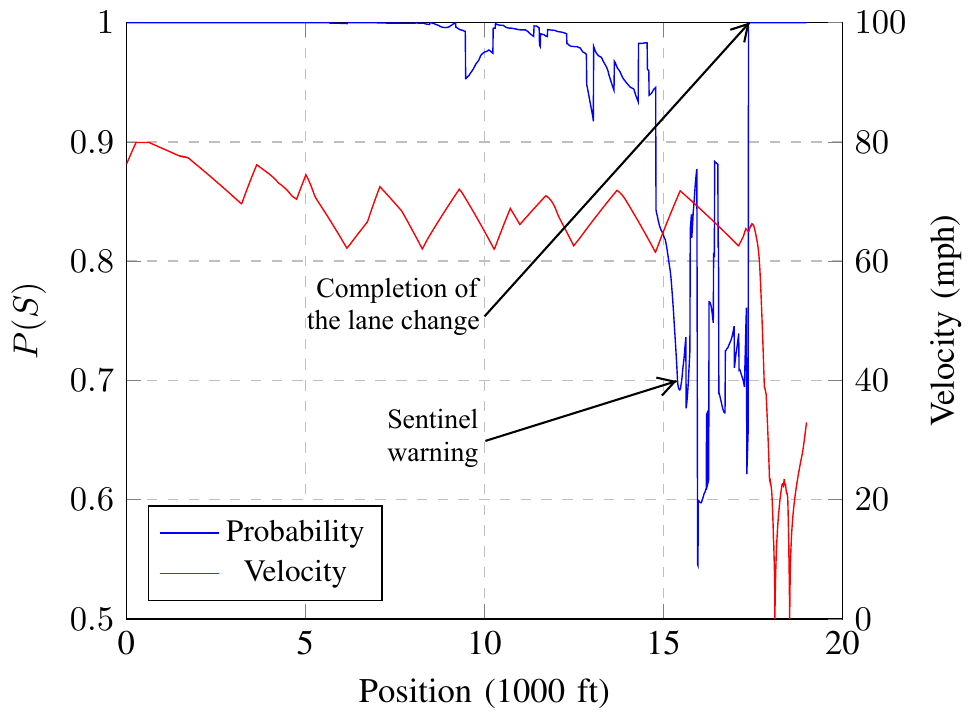}
		\caption{Estimated probability calculated by Sentinel for a sample vehicle and the vehicle's velocity for the case with $p_{l} = 0.7$. The plot shows that for the first 10,000 ft this probability is nearly 1, but as the vehicle nears the congestion boundary and vehicles in lane 2 start to slow down and get closer to each other, the probability drops until it crosses the 0.7 threshold, at which point Sentinel advises the ego vehicle to find an acceptable gap in lane 2 to leave lane 1. The vehicle does so slightly before reaching the incident-induced congestion, as indicated by a velocity profile that goes down to zero, bounces back, and does this again, portraying a stop-and-go motion.} \label{SampleProb}
	\end{figure}

\begin{figure*}[t!]
		\centering
		\includegraphics[width = 0.264\textwidth]{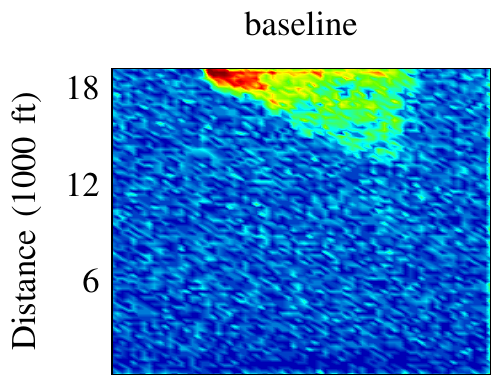} \label{L1BaselineL}
		\includegraphics[width = 0.204\textwidth]{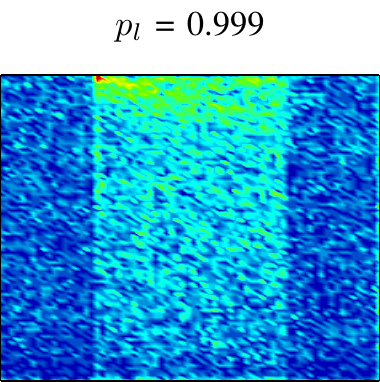} \label{L1P999L}
		\includegraphics[width = 0.204\textwidth]{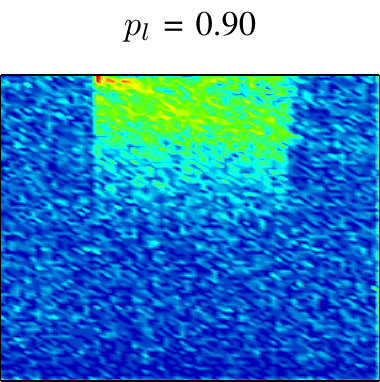} \label{LP90L}
		\includegraphics[width = 0.233\textwidth]{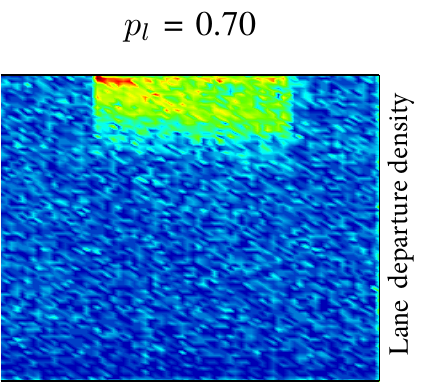} \label{L1P70L}
		\vspace*{4 pt}
		\includegraphics[width = 0.264\textwidth]{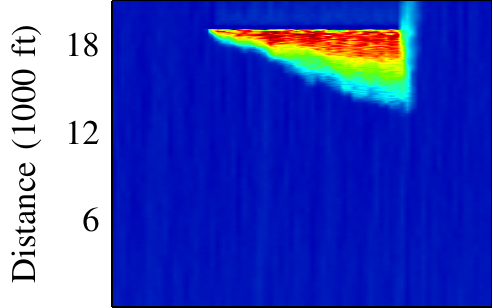} \label{L1BaselineD}
		\includegraphics[width = 0.204\textwidth]{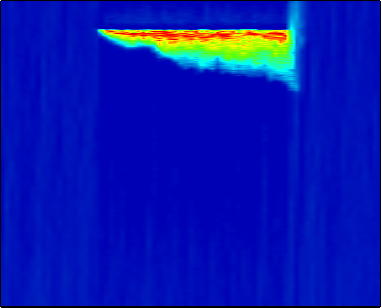} \label{L1P999D}
		\includegraphics[width = 0.204\textwidth]{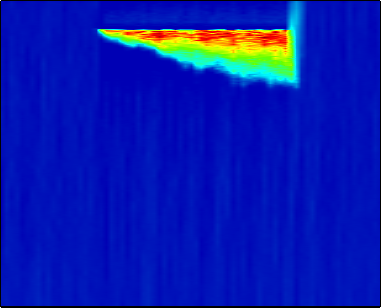} \label{L1P90D}
		\includegraphics[width = 0.233\textwidth]{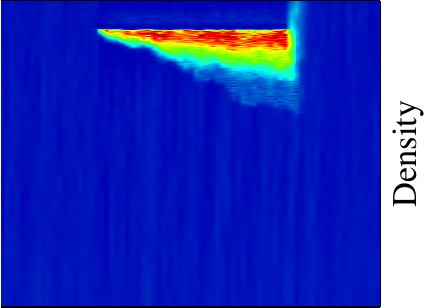} \label{L1P70D}
		\vspace*{4 pt}
		\includegraphics[width = 0.264\textwidth]{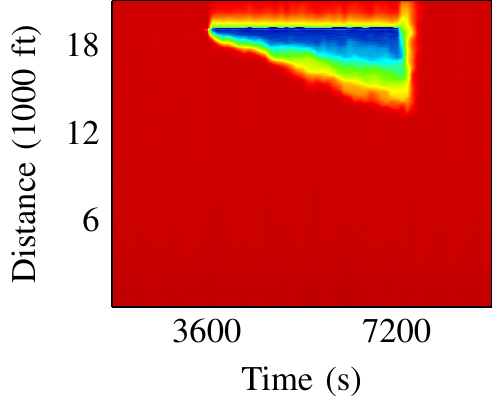} \label{L1BaselineS}
		\includegraphics[width = 0.204\textwidth]{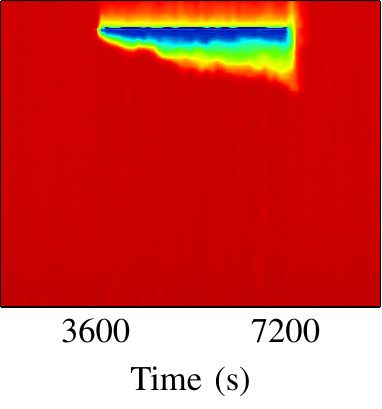} \label{L1P999S}
		\includegraphics[width = 0.204\textwidth]{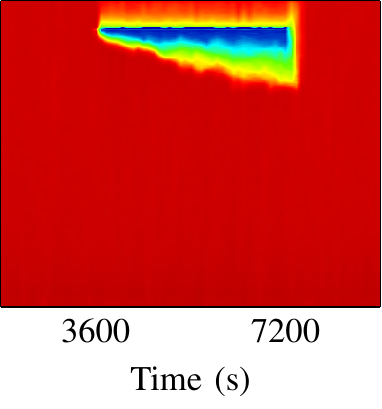} \label{L1P90S}
		\includegraphics[width = 0.233\textwidth]{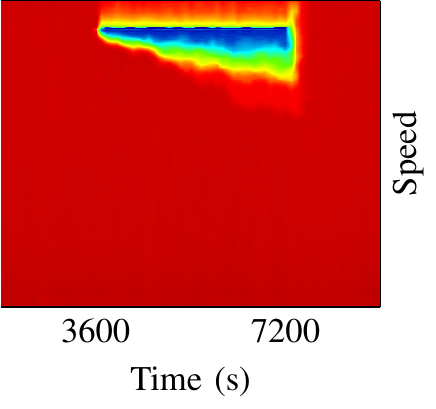} \label{L1P70S}
		\vspace*{4 pt}
		\includegraphics[width = 0.3\textwidth]{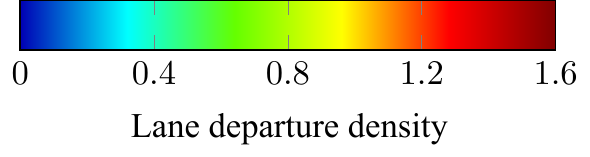} \label{L1LBar}
		\includegraphics[width = 0.3\textwidth]{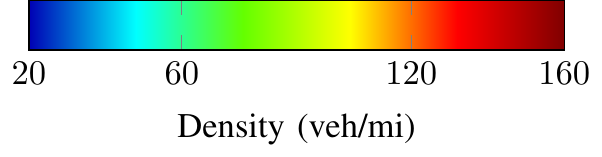} \label{L1DBar}
		\includegraphics[width = 0.3\textwidth]{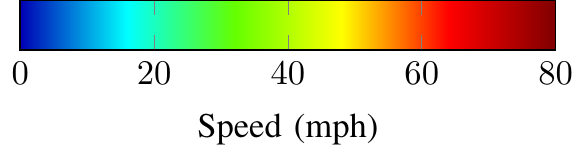} \label{L1SBar}
		\caption{Time-space plots of density and speed for all cars and lane departure density for SE cars in lane 1.} \label{L1Plots}
	\end{figure*}

\begin{figure*}[t!]
		\centering
		\includegraphics[width = 0.264\textwidth]{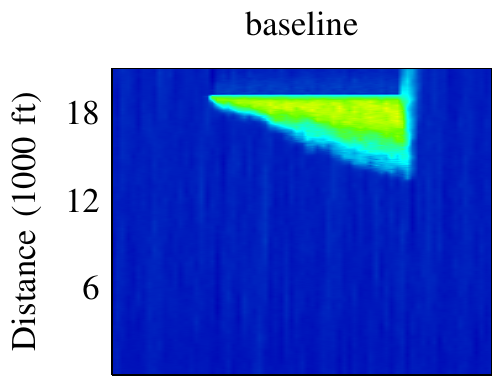} \label{L2BaselineD}
		\includegraphics[width = 0.204\textwidth]{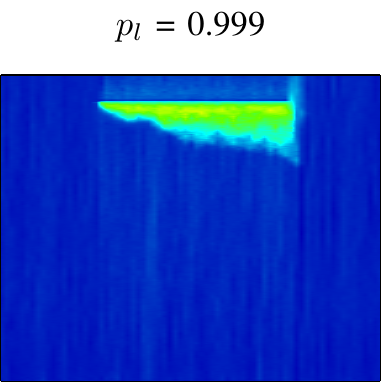} \label{L2P999D}
		\includegraphics[width = 0.204\textwidth]{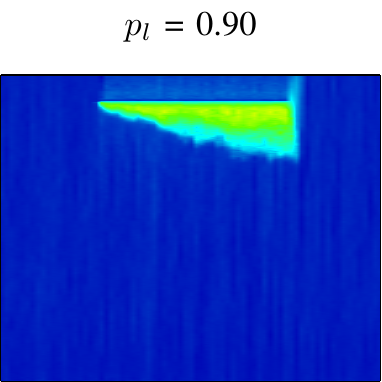} \label{L2P90D}
		\includegraphics[width = 0.232\textwidth]{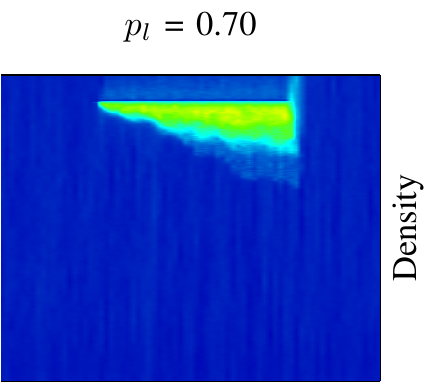} \label{L2P70D}
		\vspace*{4 pt}
		\includegraphics[width = 0.264\textwidth]{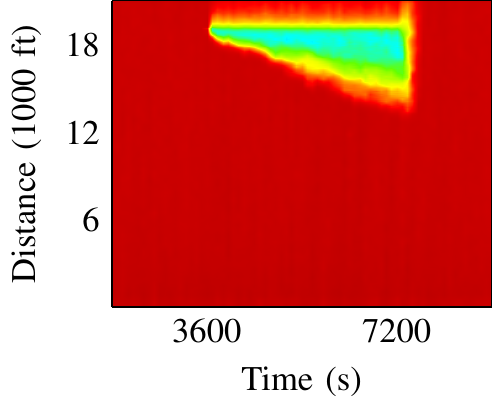} \label{L2BaselineS}
		\includegraphics[width = 0.204\textwidth]{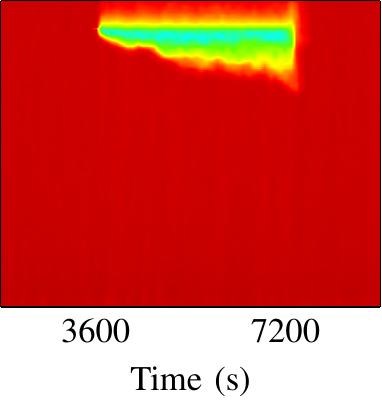} \label{L2P999S}
		\includegraphics[width = 0.204\textwidth]{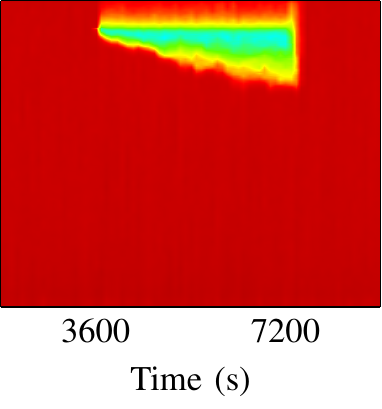} \label{L2P90S}
		\includegraphics[width = 0.233\textwidth]{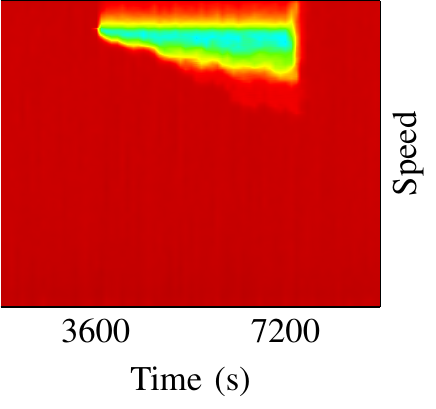} \label{L2P70S}
		\vspace*{4 pt}
		\includegraphics[width = 0.3\textwidth]{Figures/DBar.pdf} \label{L2DBar}
		\includegraphics[width = 0.3\textwidth]{Figures/SBar.pdf} \label{L2SBar}
		\caption{Time-space plots of density and speed for all vehicles in lane 2.} \label{L2Plots}
	\end{figure*}

	Let's now dive deeper into the case with $q_{i}$ = 6400 veh/hr, $r$ = 40\%, and $\gamma_{i}$ = 60 min. For this case, \autoref{Case11All} shows the average delay and discharge rate of all vehicles for each 900-second simulation time-interval during the analysis period and for different values of $p_{l}$. \autoref{Case11Regular} and \autoref{Case11Smart} show similar plots for regular cars and SE cars, respectively. Along with these figures, \autoref{SampleProb} shows the estimated $P(S)$ (probability of leaving the blocked lane before reaching the incident point) calculated by Sentinel for a sample SE car and its velocity as it travels along the road for the case with $p_{l} = 0.7$. Finally, \autoref{L1Plots} shows time-space plots of density (in veh/mi) and speed (in mph) for all vehicles and $\log(d_{l})$ for SE cars in lane 1, while \autoref{L2Plots} shows similar time-space plots of density and speed for all vehicles in lane 2. In each figure, the first column represents the baseline case, while the other three columns represent cases with $p_{l}$ = 0.999, 0.9, and 0.7, respectively. For each individual plot, the horizontal axis indicates simulation time and spans from 1800 to 9000 seconds, or the entire analysis period. The vertical axis indicates distance (in 1000 ft) and spans the entire length of the road segment (21,054 ft) for speed and density plots and up to the incident point (19,200 ft) for lane departure plots. For lane departure plots, we plotted $\log(d_{l})$ instead of $d_{l}$ to better visualize the difference in values. We also opted to only use data from SE cars for lane departure plots despite using data from all cars for density and speed plots. This was due to the fact that Sentinel only affects the behavior of SE cars and that of other cars should not be statistically different from what it was for the baseline case.
	
	\autoref{SampleProb} illustrates how Sentinel operated during the simulations. Using data from nearby vehicles obtained through perception sensors, Sentinel estimated the probability of leaving lane 1 before reaching the incident point. We can see that for the first 10,000 ft this probability was nearly 1, but as the ego vehicle neared the congestion boundary and vehicles in lane 2 started to slow down and get closer to each other, the probability dropped until it crossed the 0.7 threshold, at which point Sentinel advised the ego vehicle to find an acceptable gap in lane 2 to leave lane 1. The vehicle did so slightly before reaching the incident-induced congestion, as indicated by the probability that steps up to 1 and the velocity that goes down to zero, bounces back, and does this again, portraying a stop-and-go motion. \par
	
	This knowledge of how Sentinel operates guides our understanding of its impact on the overall traffic flow. As \autoref{Case11Smart} shows, at the fifth 900-second time interval (the first time interval after the incident), Sentinel helped SE cars leave lane 1 early, reducing their average delay relative to the baseline case. \autoref{Case11Regular} indicates that this was not the case for regular cars, whose average delay is nearly the same as the baseline case in the fifth interval. \autoref{Case11Smart} further shows that this trend of reduction in average delay continues in subsequent time intervals for SE cars, where compared to average delay's linear growth for the baseline case, the rate of growth is slower than linear in cases where Sentinel was operational. As can be seen from \autoref{Case11All} and \autoref{Case11Regular}, this behavior reduced congestion and indirectly improved average delay for other vehicles relative to the baseline case. \autoref{Case11All} also shows that Sentinel was able to restore some of the bottleneck capacity lost to congestion. It indicates that bottleneck capacity with operational Sentinel stands at around 6200 veh/hr, near the nominal capacity of a three-lane freeway and nearly 100 veh/hr more than the baseline case, with both SE cars and regular cars equally responsible for the increased capacity, as shown in \autoref{Case11Regular} and \autoref{Case11Smart}. \par
	
	\autoref{L1Plots} and \autoref{L2Plots} bring another perspective to understanding how Sentinel impacts traffic flow. The lane departure density plot for the baseline case shows that when the incident occurs (intersection of 3600 s and 19,200 ft), there is a large spike in lane departures near the incident point, which gradually dilutes along the road as congestion in lane 1 grows. In contrast, in the other three cases there is a much smaller spike at the point of incident and lane departures are more evenly distributed upstream of the incident point. Not only did this help reduce traffic density on lane 1, but it also reduced traffic density on lane 2. This was because as SE cars departed lane 1 earlier and entered lane 2, some cars on lane 2 moved to lanes 3 and 4 to maintain the balance of density on the three lanes. For the baseline case on the other hand, a large volume of cars entered lane 2 at the incident point, rapidly increasing lane 2's density because there was not enough time for other cars to move to lanes 3 and 4. \par
	
	Lane departure density plots of \autoref{L1Plots} also reveal how the choice of $p_{l}$ impacts lane departure timings. As mentioned previously, when $p_{l}$ is larger Sentinel advises vehicles to depart lane 1 earlier than it would if $p_{l}$ was lower. We can see that for $p_{l}$ = 0.999, lane departures are evenly distributed along the road. This is because any minor disturbance in the traffic caused the probability to drop below $p_{l}$ and let Sentinel advise the vehicle to depart lane 1, even if it was not anywhere near the congestion. In contrast, lane departure plots for the other two cases show that Sentinel only advised vehicles to change lanes when they got close to the congestion. For these cases, the plots show that Sentinel started advising vehicles to depart lane 1 about 1.5 mi and 0.8 mi ahead of the incident point for cases with $p_{l}$ = 0.9 and 0.7, respectively. Despite this, \autoref{60Min} shows that among the three $p_{l}$ values, the largest reduction in average delay for this traffic setting was obtained for $p_{l}$ = 0.999 followed by $p_{l}$ = 0.9. \par
	
	Overall, the choice of the optimal $p_{l}$ depends on traffic flow, road settings (number of lanes and number of blocked lanes), incident duration, and penetration rate of Sentinel. However, given that larger values of $p_{l}$ tend to push lane departures away from the incident point and distribute them more evenly, we recommend a $p_{l}$ value in the range of 0.9 to 0.97 for a satisfying performance. This value is neither too large (such as 0.999) to randomly advise vehicles to depart a lane, nor is it too small to advise vehicles to depart a lane just before reaching a congestion. In more complex cases, determining the optimal choice of $p_{l}$ may require additional traffic simulation. Dynamically assigning $p_{l}$ may be another solution, though this strategy needs further research.
	
	\section{Conclusions} \label{Conclusions}
	
	This paper introduced Sentinel, an onboard system for intelligent vehicles that guides their lane changing behavior during a freeway incident with the goal of reducing traffic congestion, capacity drop, and delay. Sentinel does so by leveraging onboard perception data, knowledge of the incident location, and a prediction model that uses traffic- and driver-related parameters to estimate the probability of reaching a goal state on the road using one or multiple lane changes. When an incident blocking the lane(s) ahead is detected, Sentinel calculates the probability of leaving the blocked lane(s) before reaching the incident point at each time step. It advises the vehicle to leave the blocked lane(s) when that probability drops below a certain threshold, as the vehicle nears the congestion boundary. By doing this, Sentinel reduces the number of disruptive, late-stage lane changes of vehicles in the blocked lane(s) trying to move to other lanes and distributes those maneuvers upstream of the incident point. \par
	
	A simulation case study in which one lane of a four-lane freeway was blocked by an incident for a set period of time was conducted using VISSIM\textsuperscript{\textregistered} to understand how Sentinel impacted traffic flow and how different parameters - traffic flow, penetration rate, and incident duration - affected Sentinel's performance. The results showed that Sentinel had a positive impact on the overall traffic flow, particularly when it had a considerable penetration rate. It was successful at reducing traffic delay by an average of 10\% or more and reduced the drop in capacity at the bottleneck. We also observed that in most cases Sentinel's performance was not strongly correlated with the choice of the probability threshold, recommending a probability threshold in the range of 0.9 to 0.97 for a satisfactory performance. \par
	
	Overall, Sentinel's effectiveness at reducing delay during freeway incidents, its simple design, and its ease of implementation show potential for its widespread adoption by intelligent vehicles, in which case it could potentially save billions of dollars annually in costs associated with congestion caused by freeway incidents. Based on the results of this study, future work will focus on using full-cabin driving simulators and real-world experiments to study driver compliance, Sentinel's performance in the real world, and its overall impact on driving behavior.
	
	\section*{Acknowledgment} \label{Section5}
	
	The authors wish to express their gratitude to Dr. Harpreet S. Dhillon for his help with the probability model and to Dr. Montasir Abbas and Awad Abdelhalim for their assistance with VISSIM\textsuperscript{\textregistered} simulations.
	
	\IEEEtriggeratref{43}
	\bibliography{FinalFormIncidentAdvisory}
	\bibliographystyle{IEEEtran}
	
	\begin{IEEEbiography}[{\includegraphics[width = 1 in, height = 1.25 in, clip, keepaspectratio]{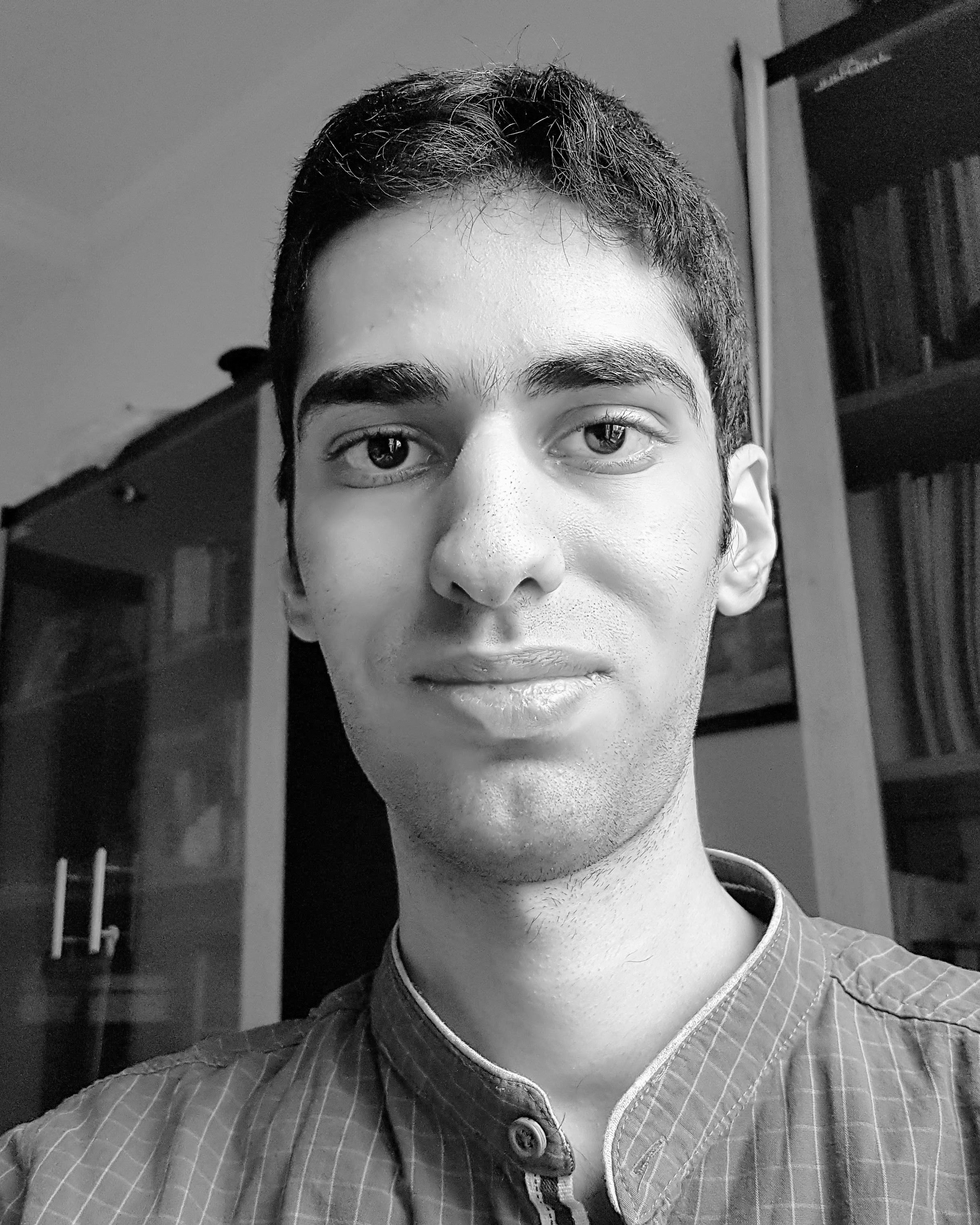}}]{Goodarz Mehr}
	
		received his B.Sc. degree in mechanical engineering from Sharif University of Technology, Tehran, Iran, and is currently pursuing a Ph.D. degree in mechanical engineering from Virginia Tech. His research interests include robotics and control, stochastic planning models, machine learning, and cooperative perception.
		
	\end{IEEEbiography}	
	\begin{IEEEbiography}[{\includegraphics[width = 1 in, height = 1.25 in, clip, keepaspectratio]{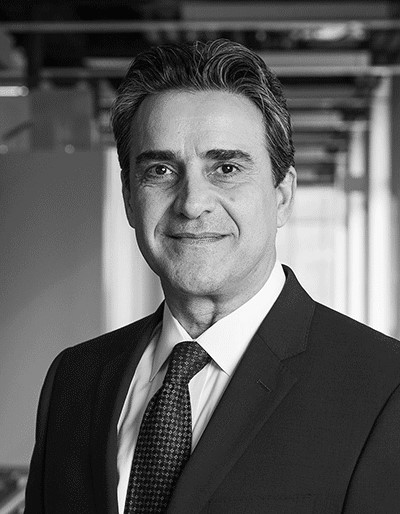}}]{Azim Eskandarian}
	
		received his B.S. degree from George Washington University (GWU), his M.S. degree from Virginia Tech, and his D.Sc. degree from GWU, all in mechanical engineering. He was a Professor of Engineering and Applied Science with GWU and the Founding Director of Center for Intelligent Systems Research from 1996 to 2015, Director of the Transportation Safety and Security University Area of Excellence from 2002 to 2015, Co-Founder of the National Crash Analysis Center in 1992, and Director of the National Crash Analysis Center from 1998 to 2002 and 2013 to 2015. He was an Assistant Professor with Pennsylvania State University, York, PA, USA, from 1989 to 1992, and an Engineer/Project Manager in industry from 1983 to 1989. He has been a Professor and the Head of the Mechanical Engineering Department at Virginia Tech (VT), since 2015. He became the Nicholas and Rebecca Des Champs Chaired Professor in 2018. He established the Autonomous Systems and Intelligent Machines Laboratory at VT to conduct research on intelligent and autonomous vehicles and mobile robots. Dr. Eskandarian is a Fellow of ASME and a member of SAE professional societies. He is a Senior Member of IEEE and received the IEEE ITS Society’s Outstanding Researcher Award in 2017 and GWU’s School of Engineering Outstanding Researcher Award in 2013.
		
	\end{IEEEbiography}
	
	\vfill
	
\end{document}